\documentclass[runningheads]{llncs}
\usepackage[T1]{fontenc}
\usepackage{graphicx,verbatim}
\usepackage{amsmath}
\usepackage{multirow}
\begin{document}
\title{RT-GAN: Recurrent Temporal GAN for Adding Lightweight Temporal Consistency to Frame-Based Domain Translation Approaches}
%

\author{Shawn Mathew\inst{1}* \and Saad Nadeem\inst{2}* \and Alvin C. Goh\inst{2} \and Arie Kaufman\inst{1}}  
\authorrunning{Mathew et al.}
\institute{Stony Brook University, New York, USA \and Memorial Sloan Kettering Cancer Center, New York, USA \\ *Equal Contribution. Corresponding Author: \url{nadeems@mskcc.org}}

\maketitle              
\begin{abstract}
Fourteen million colonoscopies are performed annually just in the U.S. However, the videos from these colonoscopies are not saved due to storage constraints (each video from a high-definition colonoscope camera can be in tens of gigabytes). Instead, a few relevant individual frames are saved for documentation/reporting purposes and these are the frames on which most current colonoscopy AI models are trained on. While developing new unsupervised domain translation methods for colonoscopy (e.g. to translate between real optical and virtual/CT colonoscopy), it is thus typical to start with approaches that initially work for individual frames without temporal consistency. Once an individual-frame model has been finalized, additional contiguous frames are added with a modified deep learning architecture to train a new model from scratch for temporal consistency. This transition to temporally-consistent deep learning models, however, requires significantly more computational and memory resources for training. In this paper, we present a lightweight solution with a tunable temporal parameter, RT-GAN (Recurrent Temporal GAN), for adding temporal consistency to individual frame-based approaches that reduces training requirements by a factor of 5. We demonstrate the effectiveness of our approach on two challenging use cases in colonoscopy: haustral fold segmentation (indicative of missed surface) and realistic colonoscopy simulator video generation. We also release a first-of-its kind temporal dataset for colonoscopy for the above use cases. The datasets, accompanying code, and pretrained models will be made available on our Computational Endoscopy Platform GitHub (\url{https://github.com/nadeemlab/CEP}). The supplementary video is available at \url{https://youtu.be/UMVP-uIXwWk}.
\keywords{Temporal GAN \and Colonoscopy \and Domain Translation}
\end{abstract}
\section{Introduction}
More than 14 million colonoscopies are performed every year, just in the U.S. Even though there is a LIVE video feed guiding the navigation of the enodoscopist during the 20--40 min minimally-invasive procedure, hardly any of these videos are stored for later analysis due to storage constraints/costs (each high-definition video can be several gigabytes); instead a few relevant frames are stored for reporting/documentation purposes only. To help assist endoscopists during procedures for tumor detection or to document the quality of the procedure for education purposes (e.g. by tracking the colon surface area missed during procedure via haustral fold occlusion -- higher surface area missed equates to higher possibility of missed cancer), AI models are normally trained on the few stored frames easily accessible via electronic health records (through Institutional Review Board approval). \textit{Rather than trying to train video models from scratch with vast amounts of video data (NOT available), can we add lightweight temporal consistency to our best-performing single-frame models for video analysis to aid endoscopists?} We address this question in this paper.


Recently, unsupervised domain translation methods have shown promising results across different colonoscopy tasks (e.g. to translate between optical [OC] and prior-treatment virtual/CT colonoscopy [VC]), but not all have been extended to video. The domain translation models that have been extended to video create new models from scratch to accommodate video sequences. Once a frame-based model has been finalized, one can either try simple post-processing normalization across frames to get ``quasi-consistency'' \cite{mathew2020augmenting} or train a new model from scratch with full temporal consistency. The first approach is only possible on very specific tasks, such as depth estimation, where there is one correct result. Tasks such as realistic image generation cannot be concatenated together with simple approaches (results will flicker as shown in \textbf{supplementary video}). The second, more general option however requires significantly more computational and memory resources for training. Moreover, temporally-consistent unsupervised video-to-video domain translation (RecycleGAN \cite{bansal2018recycle} derivatives) typically requires learning both directions of translation when only a single direction may be relevant, for example, colonoscopy to depth, colonoscopy to fold segmentation, synthetic to real colonoscopy simulation, etc. This forward and backward learning with temporal components increases the number of learnable parameters by several orders of magnitude. Even still, the general approaches like RecycleGAN may not utilize domain specific knowledge that can vastly improve results. Incorporating domain specific contributions from frame based models can be quite involved and time consuming.

\begin{figure*}[t]
\begin{center}
\setlength{\tabcolsep}{1pt}
\begin{tabular}{c|c|c}

\includegraphics[width=0.24\textwidth]{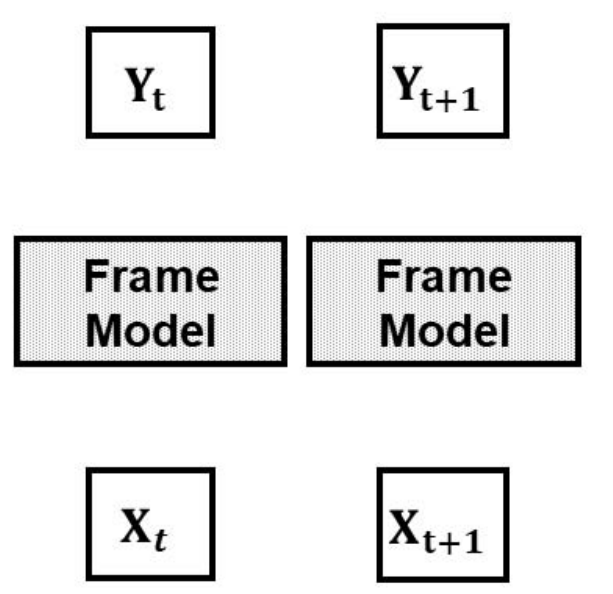}&
\includegraphics[width=0.32\textwidth]{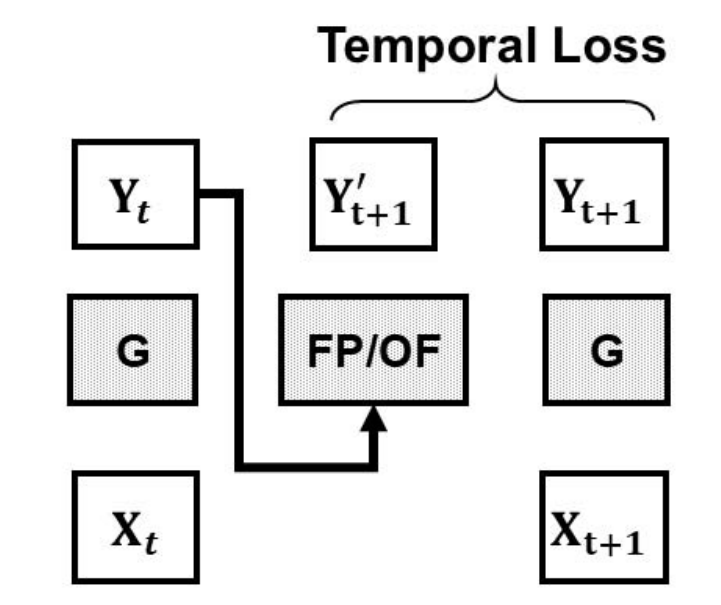}&
\includegraphics[width=0.42\textwidth]{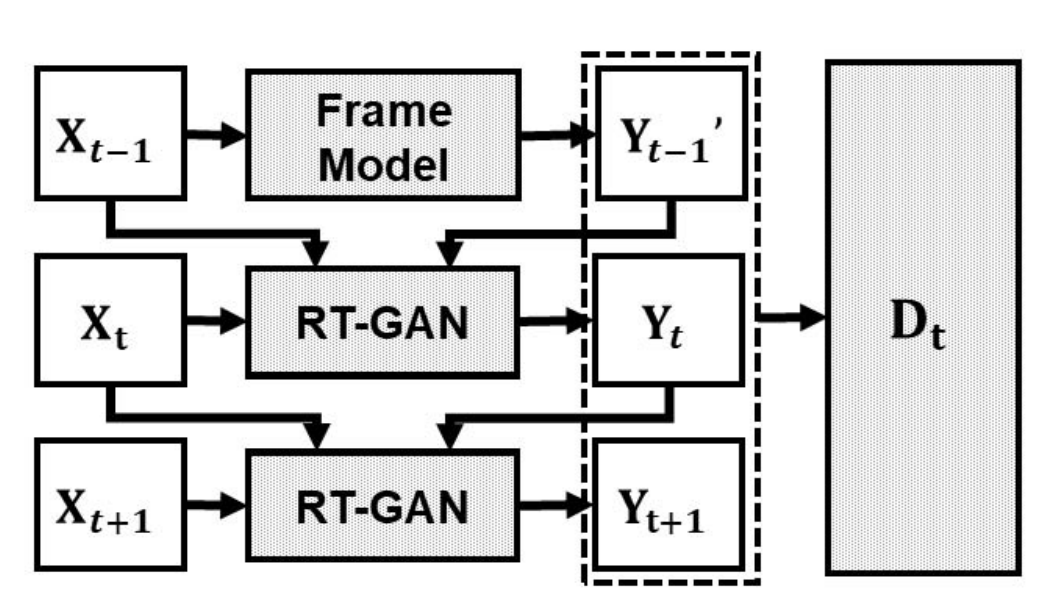}\\
(a) Frame model & (b) Temporal model & (c) RT-GAN
\end{tabular}
\caption{Depicting how temporal consistency can be added. X is the input video and Y is the resulting video output. (a) Frame-based model, (b) Temporal consistency is added in RecycleGAN \cite{bansal2018recycle} and OfGAN\cite{xu2020ofgan} using optical flow or future frame prediction. (c) RT-GAN uses three consecutive output frames from generators are passed into a discriminator to provide temporal consistency. The first frame, $Y_{t-1}'$ is generated from a fully trained frame-based model. The other two frames are created by RT-GAN to be temporally consistent with $Y_{t-1}'$.}
\label{fig:overview}
\end{center}
\end{figure*}

In this work, we present Recurrent Temporal Generative-Adversarial Network (RT-GAN) for adding lightweight temporal-consistency to unsupervised image-to-image domain translation models (that reduces training requirements by a factor of 5). RT-GAN allows traversal between temporal consistency and fidelity to the frame-based models using a single tunable weight parameter while focusing on a single translation direction. Specifically, RT-GAN uses recurrent information by referencing the previous frame and its result as seen in Figure \ref{fig:overview}c. A temporal discriminator takes the generator's results for 3 consecutive frames to build temporal consistency; using only 3 sequential frames for temporal learning was a design choice to minimize resource utilization. In essense, RT-GAN builds on the representations learned by any unsupervised image-to-image domain translation model and adds temporal consistency to these without needing to redesign task-specific components. We demonstrate the effectiveness of RT-GAN in adding lightweight temporal consistency to two frame-based models, FoldIt \cite{mathew2021foldit} haustral fold segmentation model with some inherent ``quasi-consistency'' across frames and CLTS-GAN \cite{mathew2022clts} color-lighting-texture-specular reflection augmentation model with no consistency at all across frames.\\

\noindent
\textbf{Related Work}
Bashkirova et al. performed a number of experiments using a CycleGAN model that uses 3D convolutions with varying input types such as randomly sorted frames, ordered frames, and  frames stacked as a 3D tensor. They found that using the stacking frames into 3D tensors provided the best results at the cost of extra training requirements \cite{bashkirova2018unsupervised}. Bansal et al. proposed RecycleGAN \cite{bansal2018recycle}, a network for unsupervised video retargeting, that does not require any task-specific modules and adds temporal consistency components (such as optical flow) on CycleGAN to extend it to videos (Figure \ref{fig:overview}b). Specifically, an additional future frame prediction network is added for temporal consistency. This increases memory requirements especially since two predictor networks are needed, one for each domain. OfGAN \cite{xu2020ofgan} predicts optical flow using an architecture similar to the one shown in Figure \ref{fig:overview}b to translate synthetic colonoscopy sequences to real colonoscopy video sequences; OfGAN relies on texture, lighting, and specular reflection information to be embedded in the input videos to generate realistic colored output sequences. CycleSTTN \cite{daher2023cyclesttn} is another recent work that learns a video domain translation task for specular augmentation, however, it requires paired data generation which may not be feasible for all tasks.

\section{RT-GAN: Recurrent Temporal GAN}
\noindent
\textbf{Dataset:} The OC and VC dataset was created from 10 patients at Stony Brook University Hospital that had VC procedures followed by OC procedures. The OC videos were cropped to a size of 256x256 to remove borders in the frames created by the fish-eye lens in the colonoscope. The videos for VC were created from triangulated meshes of the colon extracted from CT scans as described by Nadeem et al. \cite{nadeem2016computer}. A virtual camera flies through the mesh with random rotations and lights at both sides of the camera. To better replicate the conditions of the colonoscopy procedure, the inverse square fall-off property is applied to the lights \cite{mahmood2018unsupervised}. The videos for both the VC and OC datasets were split into 300 sets of 3 sequential frames. In total, training, validation and testing datasets are composed of 1500, 900 and 600 frame triplets respectively. Haustral fold segmentation data is generated in a similar manner to Mathew et al. \cite{mathew2021foldit}. The VC 3D meshes will be publicly released as well for video generation via Blender or VR-CAPS\cite{incetan2021vr}.

\begin{figure*}[t!]
\begin{center}
\setlength{\tabcolsep}{2pt}
\begin{tabular}{ccccc|cccc}

\rotatebox{90}{\rlap{\scriptsize ~~Input}}&
\includegraphics[width=0.11\textwidth]{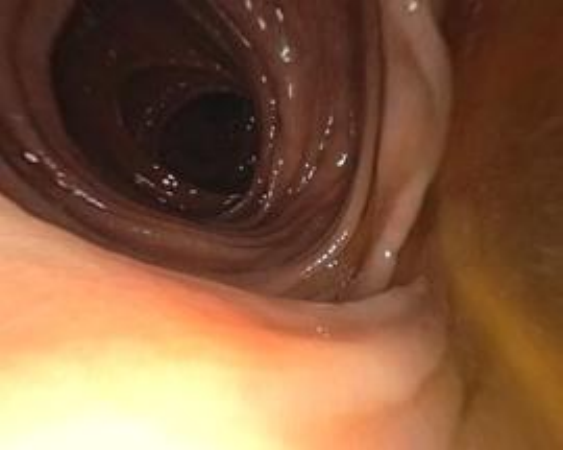}&
\includegraphics[width=0.11\textwidth]{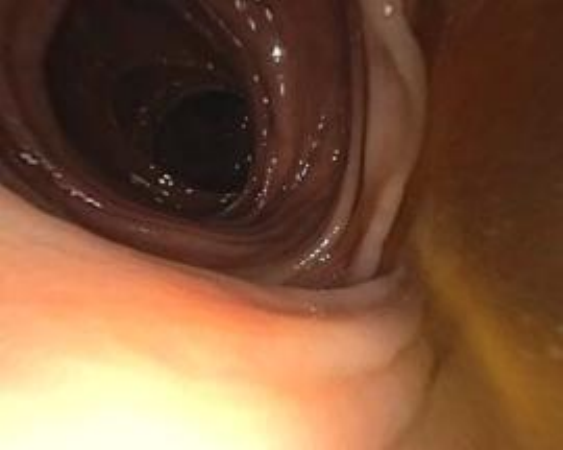}&
\includegraphics[width=0.11\textwidth]{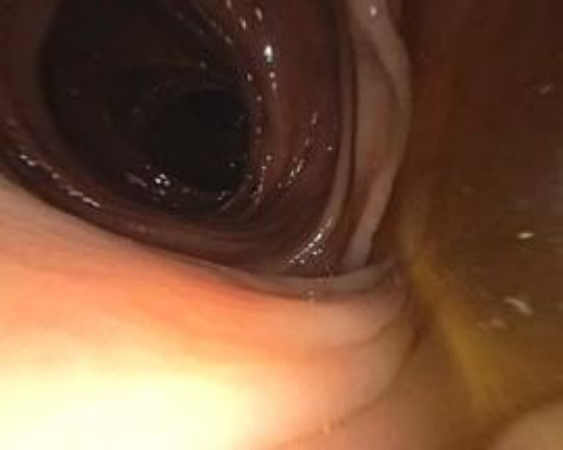}&
\includegraphics[width=0.11\textwidth]{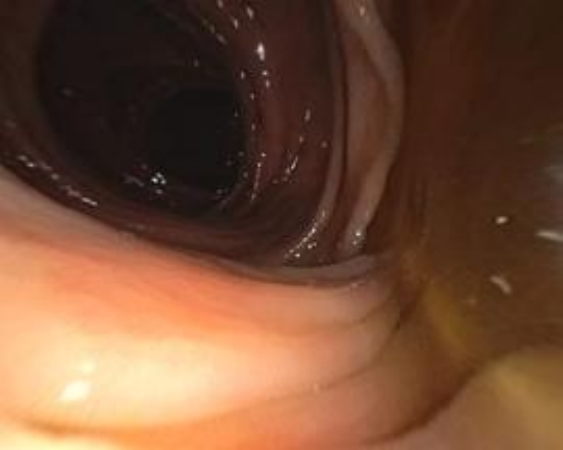}&

\includegraphics[width=0.11\textwidth]{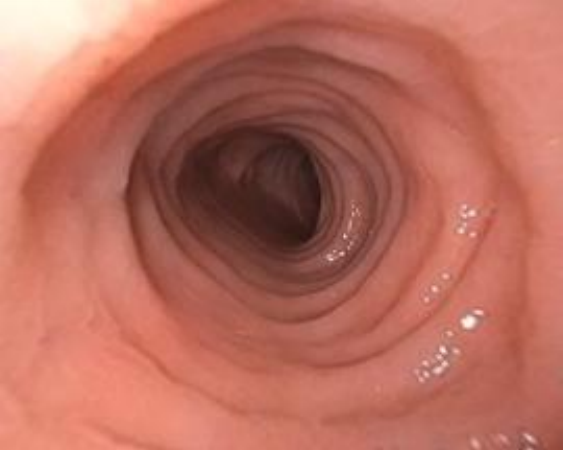}&
\includegraphics[width=0.11\textwidth]{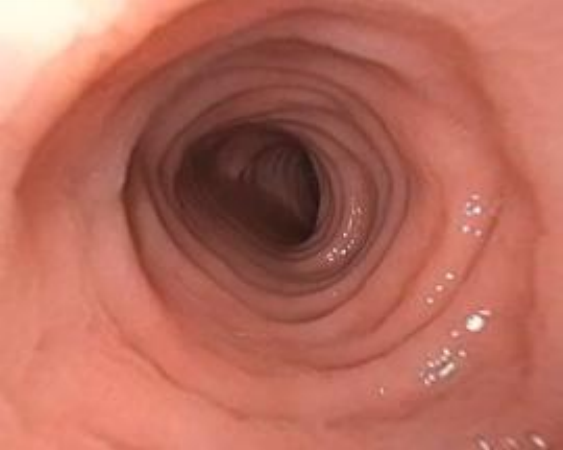}&
\includegraphics[width=0.11\textwidth]{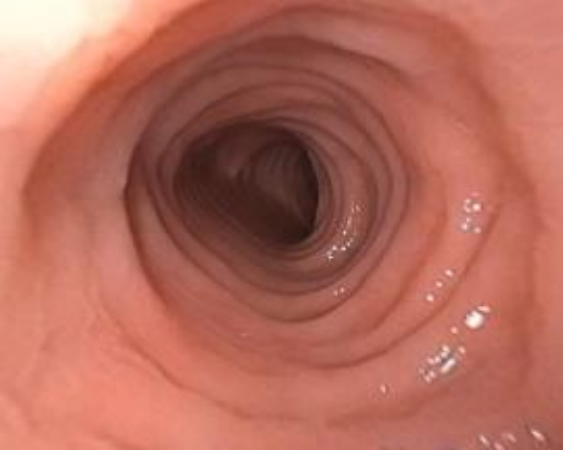}&
\includegraphics[width=0.11\textwidth]{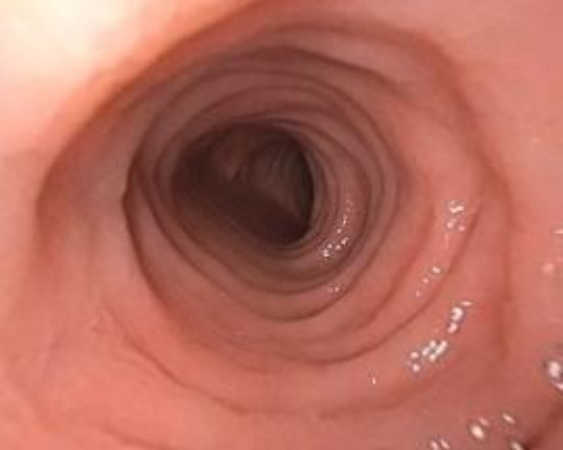}\\

\rotatebox{90}{\rlap{\scriptsize ~~FoldIt}}&
\includegraphics[width=0.11\textwidth]{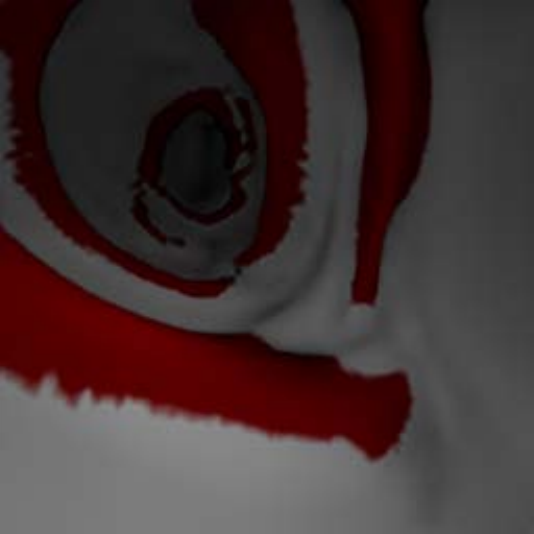}&
\includegraphics[width=0.11\textwidth]{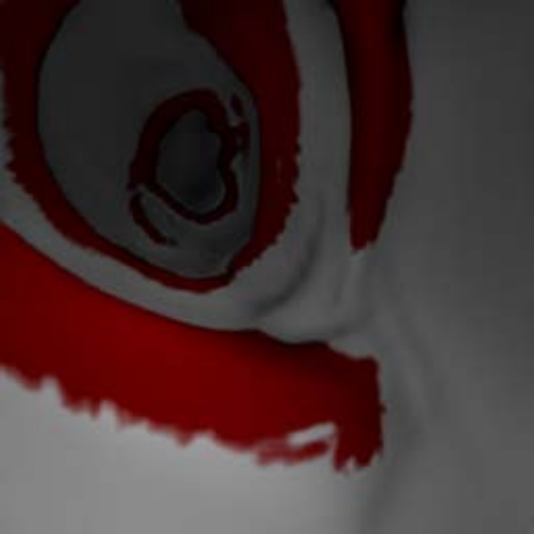}&
\includegraphics[width=0.11\textwidth]{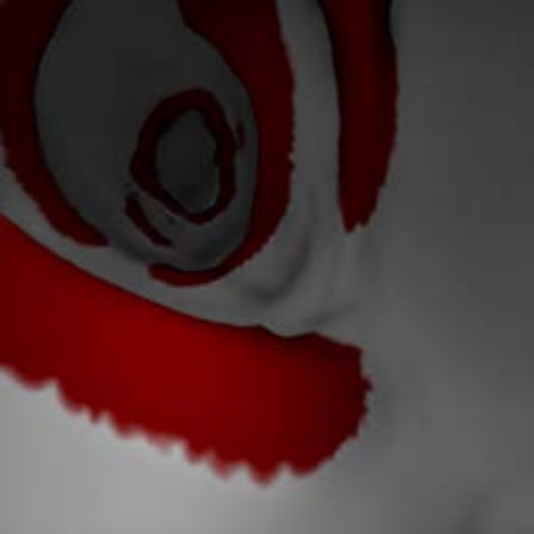}&
\includegraphics[width=0.11\textwidth]{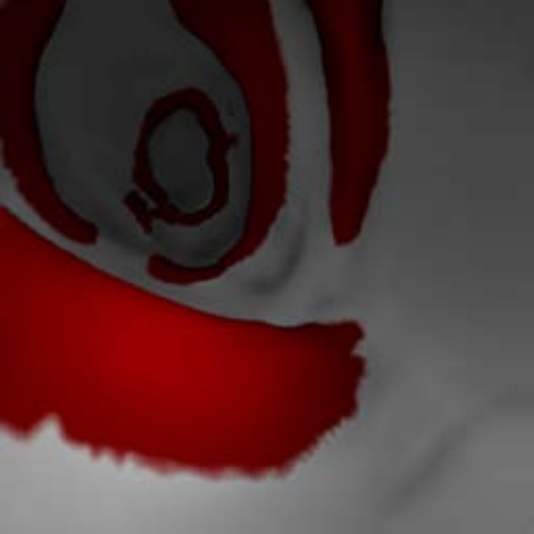}&

\includegraphics[width=0.11\textwidth]{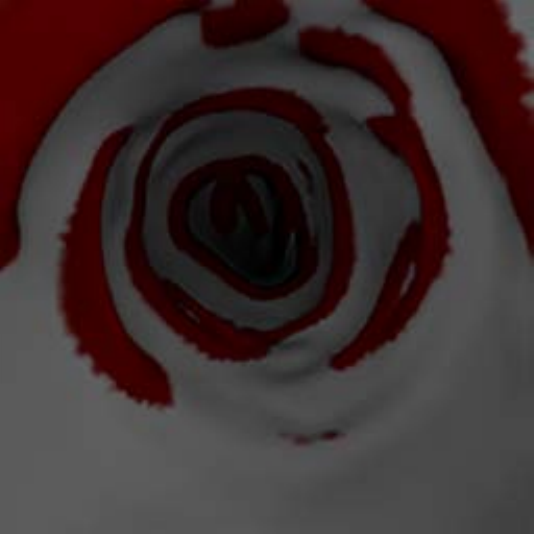}&
\includegraphics[width=0.11\textwidth]{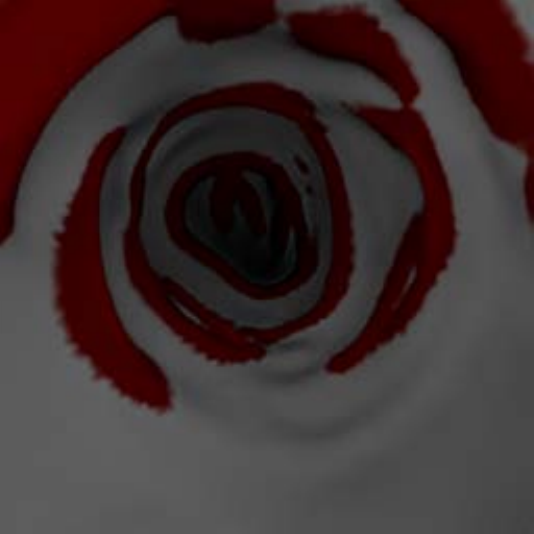}&
\includegraphics[width=0.11\textwidth]{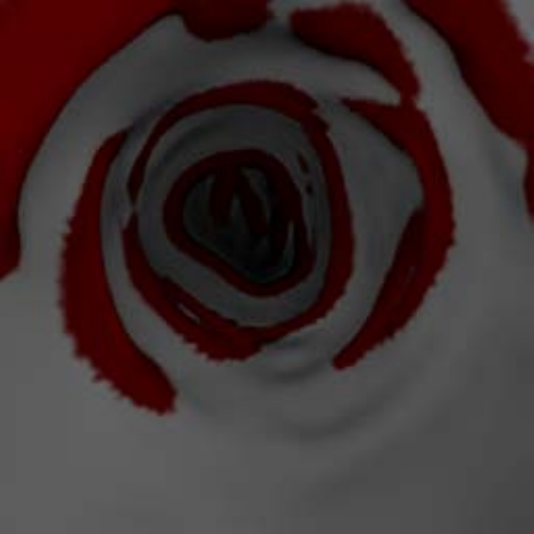}&
\includegraphics[width=0.11\textwidth]{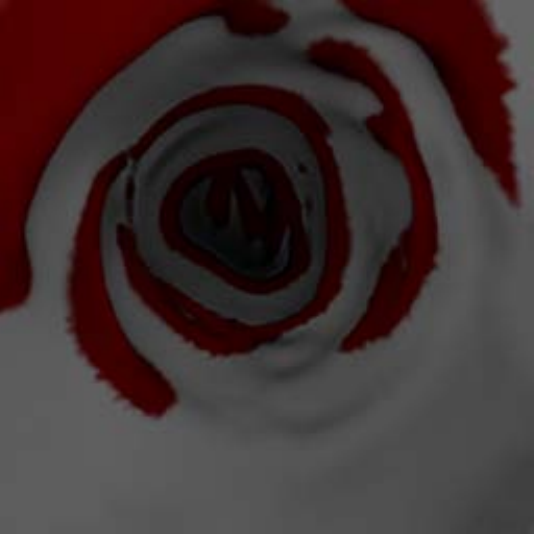}\\

\rotatebox{90}{\rlap{\tiny TempCycle}}&
\includegraphics[width=0.11\textwidth]{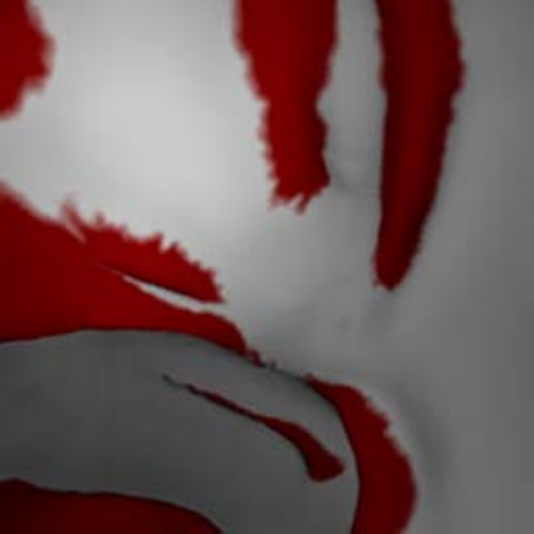}&
\includegraphics[width=0.11\textwidth]{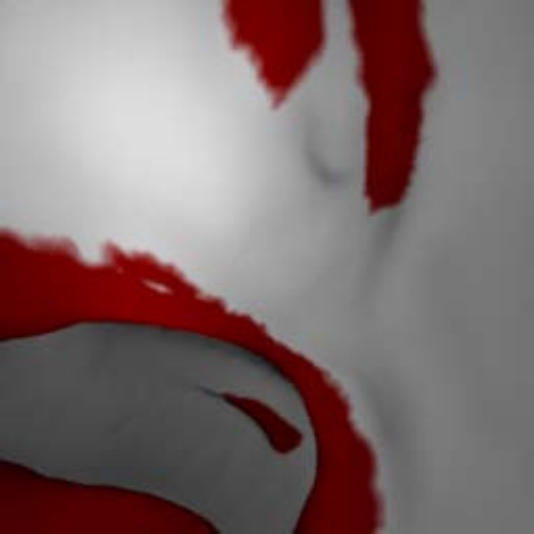}&
\includegraphics[width=0.11\textwidth]{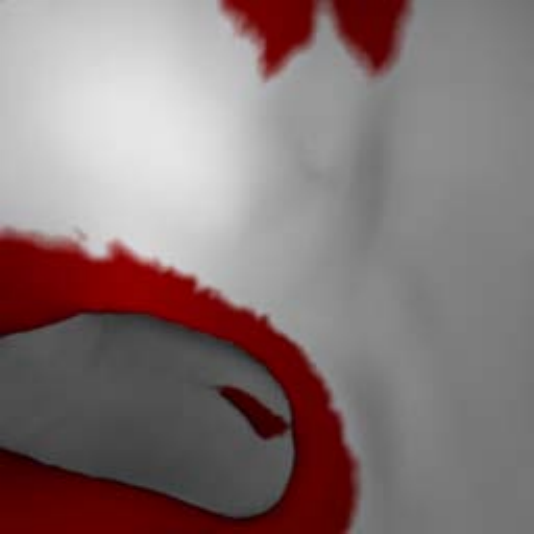}&
\includegraphics[width=0.11\textwidth]{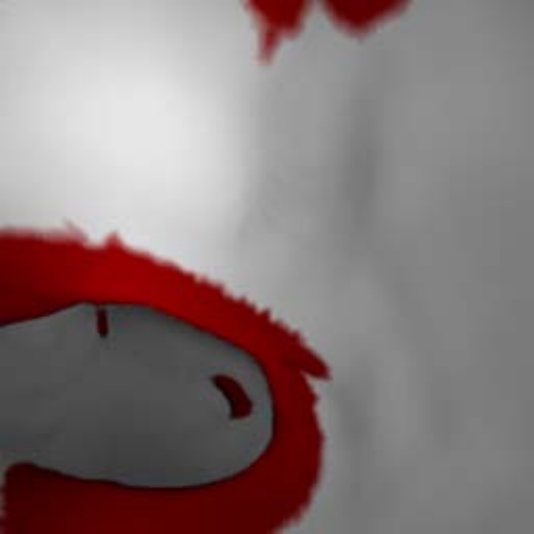}&

\includegraphics[width=0.11\textwidth]{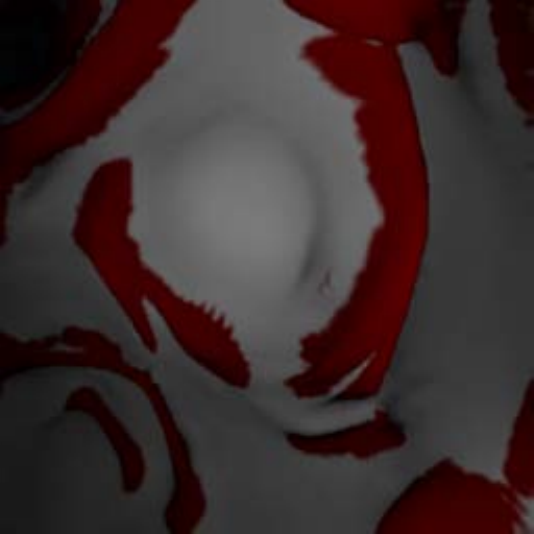}&
\includegraphics[width=0.11\textwidth]{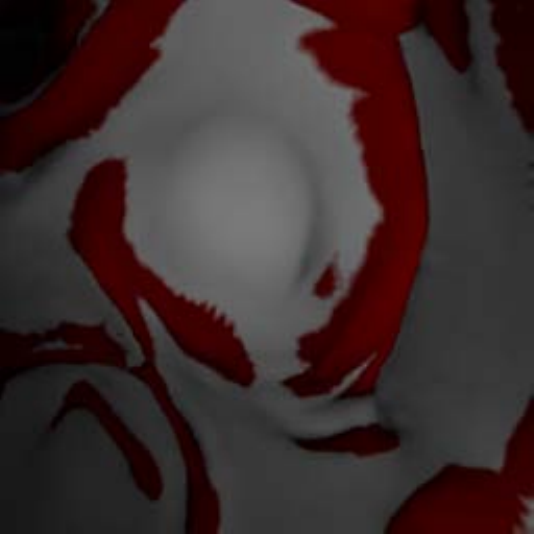}&
\includegraphics[width=0.11\textwidth]{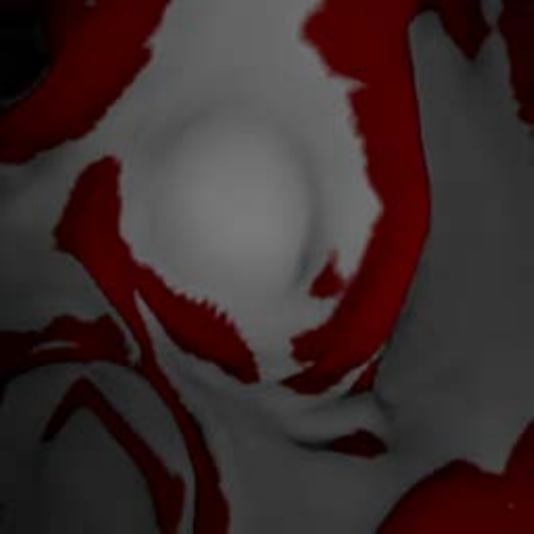}&
\includegraphics[width=0.11\textwidth]{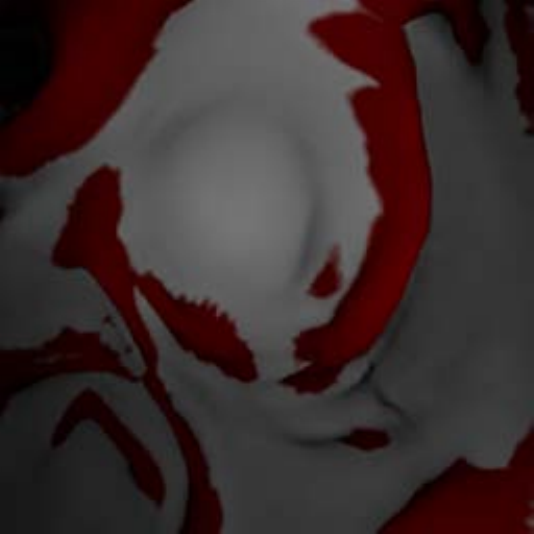}\\

\rotatebox{90}{\rlap{\tiny ~Recycle}}&
\includegraphics[width=0.11\textwidth]{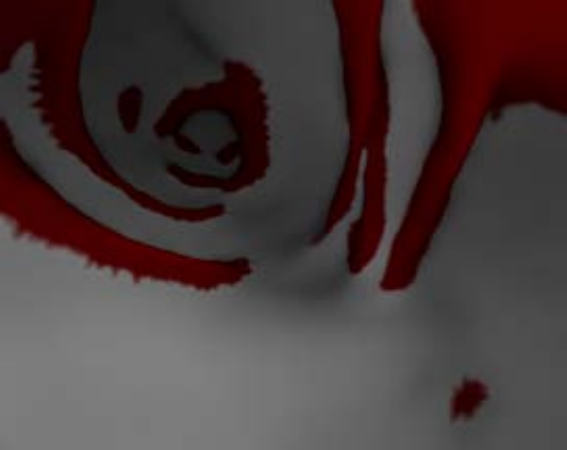}&
\includegraphics[width=0.11\textwidth]{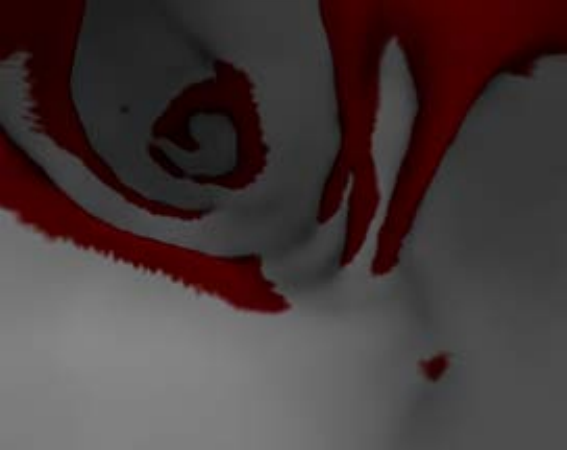}&
\includegraphics[width=0.11\textwidth]{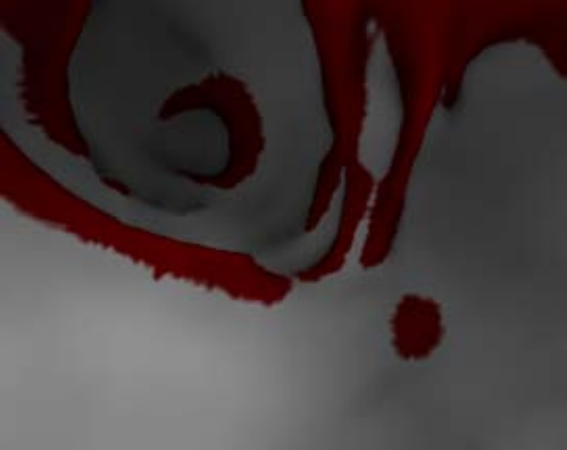}&
\includegraphics[width=0.11\textwidth]{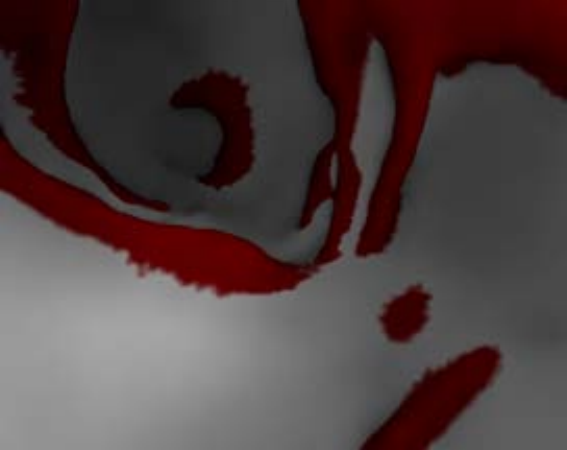}&

\includegraphics[width=0.11\textwidth]{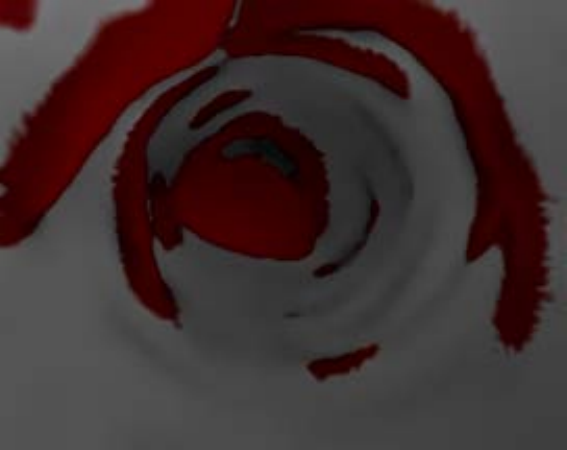}&
\includegraphics[width=0.11\textwidth]{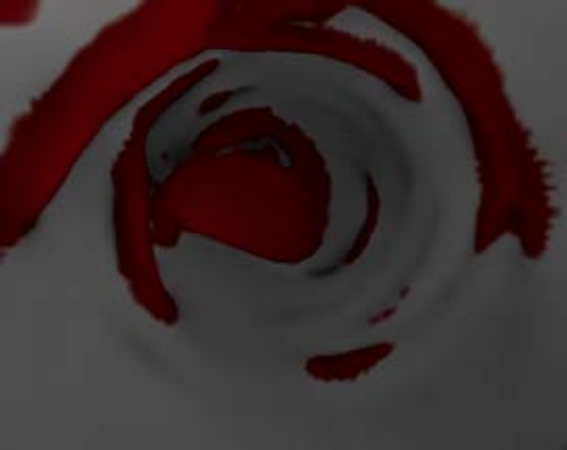}&
\includegraphics[width=0.11\textwidth]{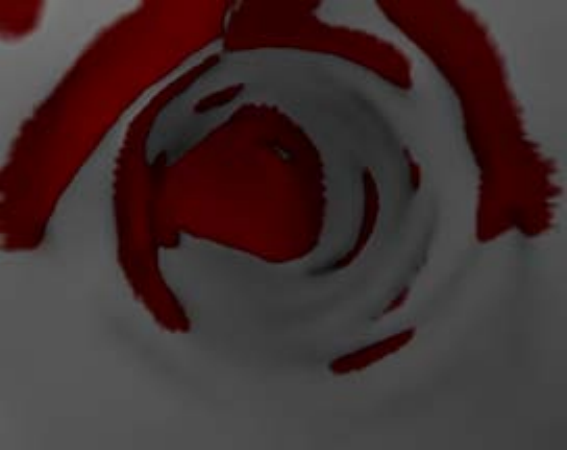}&
\includegraphics[width=0.11\textwidth]{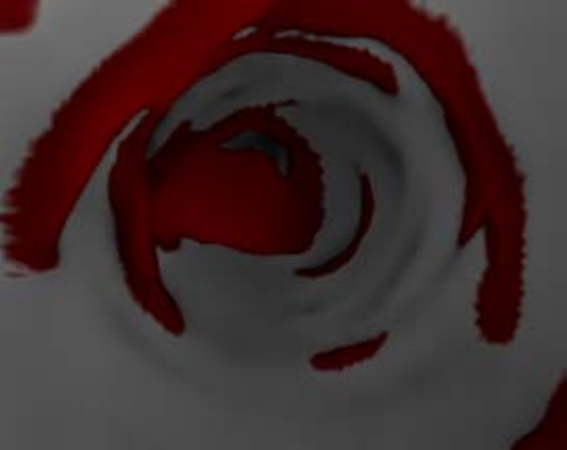}\\

\rotatebox{90}{\rlap{\scriptsize ~RT-GAN}}&
\includegraphics[width=0.11\textwidth]{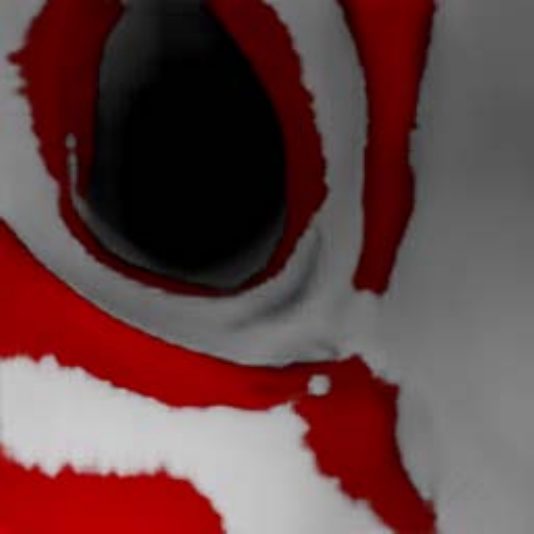}&
\includegraphics[width=0.11\textwidth]{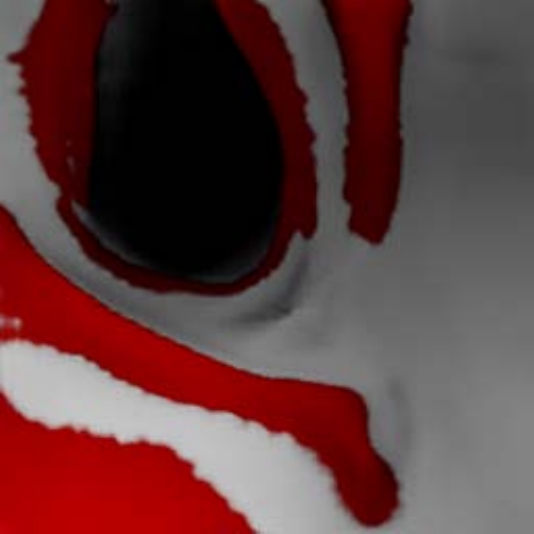}&
\includegraphics[width=0.11\textwidth]{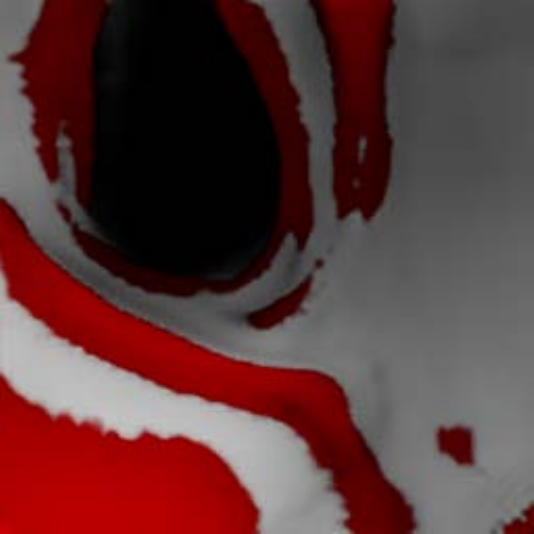}&
\includegraphics[width=0.11\textwidth]{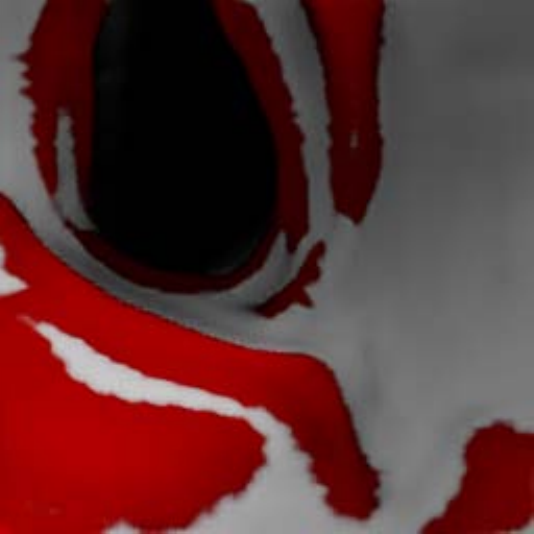}&

\includegraphics[width=0.11\textwidth]{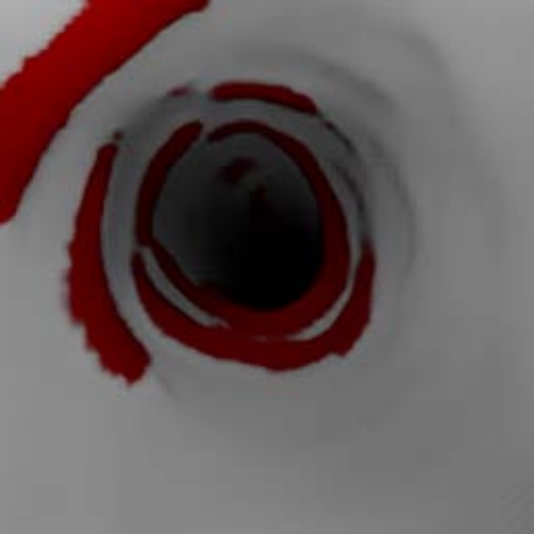}&
\includegraphics[width=0.11\textwidth]{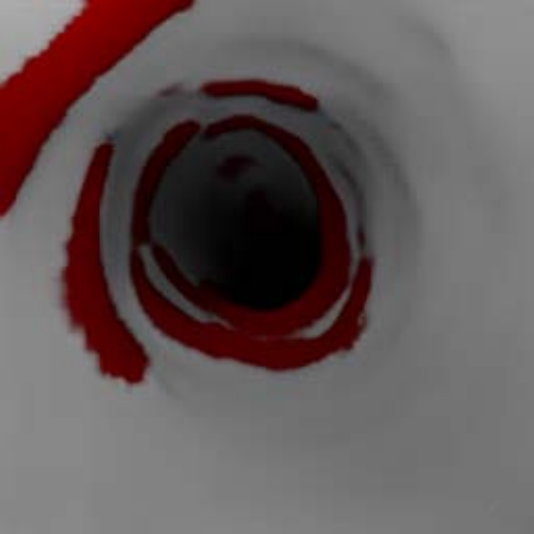}&
\includegraphics[width=0.11\textwidth]{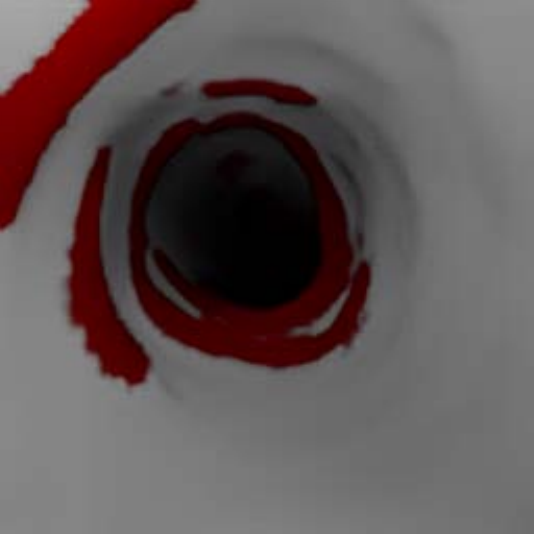}&
\includegraphics[width=0.11\textwidth]{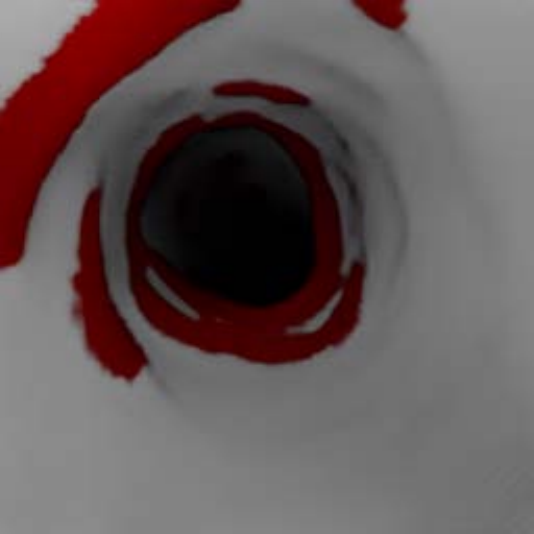}\\

\end{tabular}
\caption{Comparisons of the results for RT-GAN (Ours) with stitched images from FoldIt, TempCycleGAN, and RecycleGAN on optical colonoscopy video dataset from \cite{ma2019real}. Full results are found in the \textbf{supplementary video}.}
\label{fig:recycle_comp}
\end{center}
\end{figure*}

\noindent
\textbf{Methods:} Typically for unsupervised domain translation, at least 2 generator networks are being updated during training time. One generator learns the translation between the input domain and output domain, while the other learns the inverse direction. Typically, only one generator is required to provide the domain translation results for an application. RT-GAN only trains one generator reducing resources during training (see Table \ref{tab:params}). RT-GAN builds off of the results from a fully trained frame-based model, $F$. The results of the frame-based model can be pre-computed, so it does not affect the required resources for training. Resource requirements will be defined by the more resource hungry frame-based models. The RT-GAN's generator $G$, translates from the input domain $X$ to the output domain $Y$. $G$ takes 3 images as input to produce the output $y_t'$. The first input is the frame, $x_t$ that is to be translated.  The next input is the previous frame in the input sequence, $x_{t-1}$, to give the network context and a better understanding of motion. The last input image for $G$ is $y_{t-1}'$, the result for $x_{t-1}$. $y_{t-1}'$ gives the generator context on the previous frame with which the output needs to be temporally consistent with. The input for RT-GAN's generator can be seen in Figure \ref{fig:overview}c.

$G$ is trained using two discriminators, each having its own adversarial loss. The adversarial/discriminator loss is described below:
\begin{equation}
     \mathcal{L}_{adv}(G,D,y,y') =  \textrm{log} (D(y)) + \\
     \textrm{log}(1 - D(y')),
\end{equation}
where $y'$ is from the generator and $y$ is from the training data.

The first discriminator, $D_t$, learns temporal consistency. $D_t$ compares a 3 frame sequence from the output domain to a 3 frame sequence created from the generators. The first frame in the triplet is provided by F, while the next 2 temporally consistent frames are provided by $G$. $G$ aligns its results with $F$ in order to provide temporal consistency, but $F$'s results is indepedent of $G$. The temporal adversarial loss is described as,
\begin{equation}
\begin{aligned}
    \mathcal{L}_{t}(G,F,&D_t,Y,X)  = \\
    & \mathcal{L}_{adv}(G,D_t,\{y_{t-1},y_{t},y_{t+1} \},\{F(x_{t-1}),y_t', y_{t+1}' \}),
\end{aligned}
\end{equation}
where $y_t'$ is $G(x_{t-1},x_t,F(x_t))$ and $y_{t+1}'$ is $G(x_t,x_{t+1},y_t')$. 

A separate discriminator, $D_f$, ensures that $G$'s results appear similar to $F$. It compares the paired input and output frames for $F$ and $G$. The adversarial loss for $D_f$ is described as:
\begin{equation}
     \mathcal{L}_{f}(G,F,D_f,X) =  \mathcal{L}_{adv}(G,D_f, \{x_t,F(x_t)\},\{x_t,y_t'\}) 
\end{equation}


If the camera and the subject do not move between two frames, the resulting output of the model should also should not change. A stationary loss $\mathcal{L}_{s}$ is added for this which also helps enforce the model to use the previous frame's output ($y_t$') rather than just predicting from the current frame. The stationary loss is defined as:

\begin{equation}
     \mathcal{L}_{s}(G,X) =   \|y_t' - G(x_t,x_t,y_t')\|_1 ,
\end{equation}
where $\| \cdot \|$ represents the $\ell 1$ norm. \textit{Note that the stationary loss differs quite substantially from a perceptual loss. The stationary loss ensures temporal stability, while a perceptual loss is meant to improve the quality of the image.} 

The complete objective function for the network is:
\begin{equation}
     \mathcal{L}_{obj} = \lambda\mathcal{L}_{t}(G,F,D_t,Y,X) + \mathcal{L}_{f}(G,F,D_f,X) + \mathcal{L}_{S}(G,X)
\end{equation}
where $\lambda$ is a tunable weight to determine the tradeoff between the temporal smoothness and fidelity to the frame-based model. $D_t$ is a PatchGAN discriminator with 3D convolutions to help it learn temporal information while $D_f$ is a PatchGAN discriminator with 2D convolutions to learn spatial information. $G$ uses a Resnet architecture with 9 blocks. \textit{The bottleneck for training parameters is determined by the frame-based model. Trainable parameters is the max(frame\_model, RT-GAN) and training time is frame\_model + RT-GAN.}

\begin{figure*}[t!]
\begin{center}
\setlength{\tabcolsep}{2pt}
\begin{tabular}{ccccccccc}

\rotatebox{90}{\rlap{\small ~~Input}}&
\includegraphics[width=0.11\textwidth]{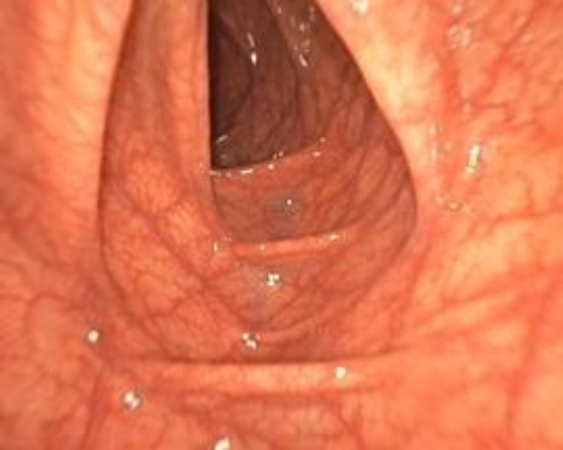}&
\includegraphics[width=0.11\textwidth]{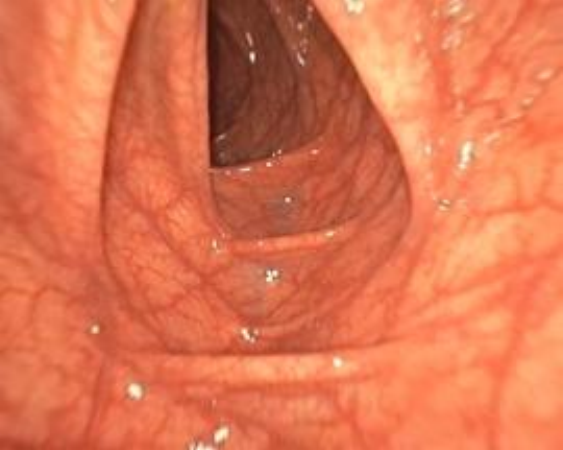}&
\includegraphics[width=0.11\textwidth]{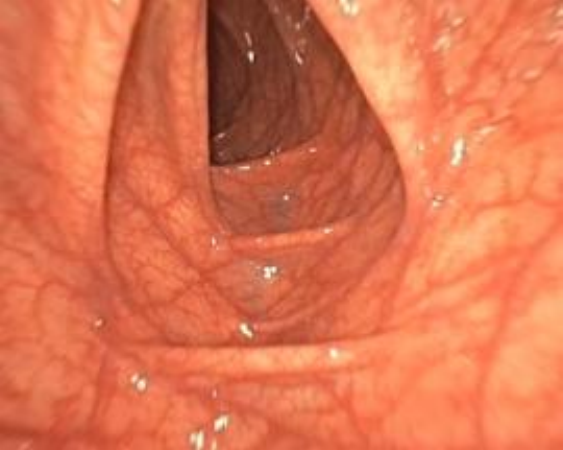}&
\includegraphics[width=0.11\textwidth]{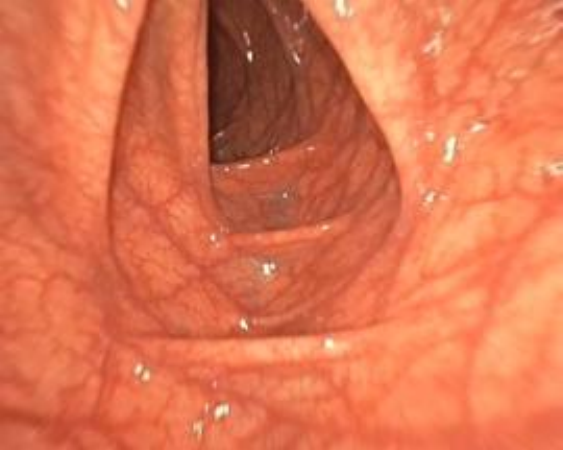}&
\includegraphics[width=0.11\textwidth]{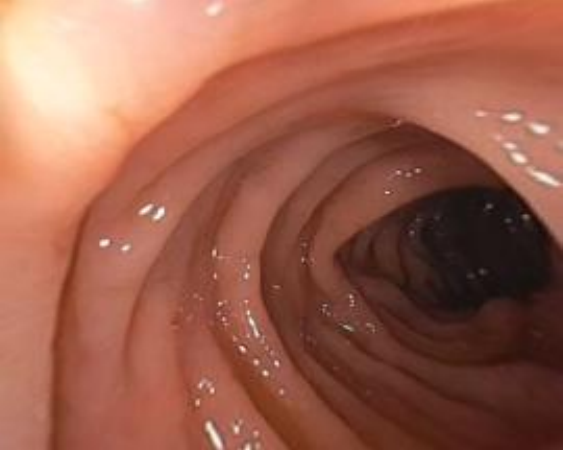}&
\includegraphics[width=0.11\textwidth]{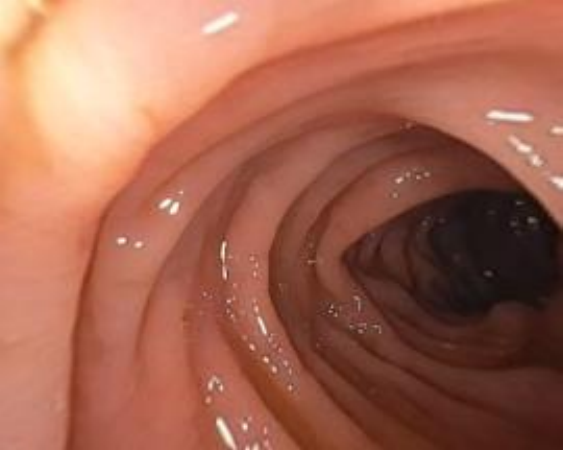}&
\includegraphics[width=0.11\textwidth]{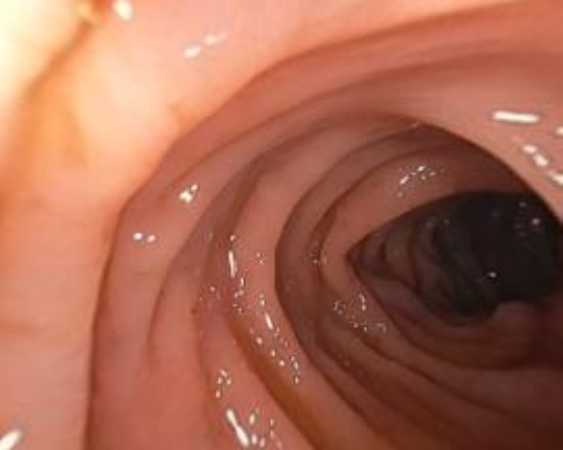}&
\includegraphics[width=0.11\textwidth]{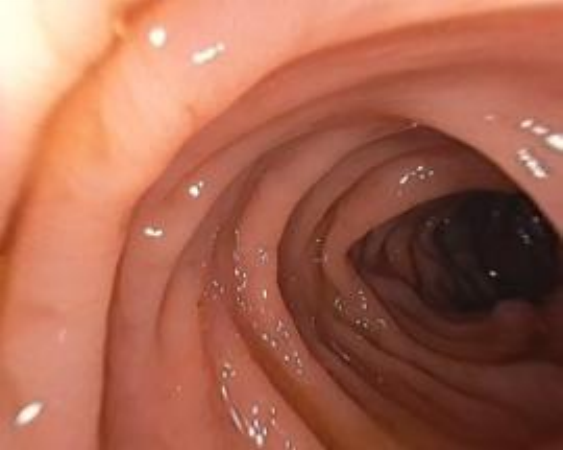}\\

\rotatebox{90}{\rlap{\small ~~$\lambda = 0.2$}}&
\includegraphics[width=0.11\textwidth]{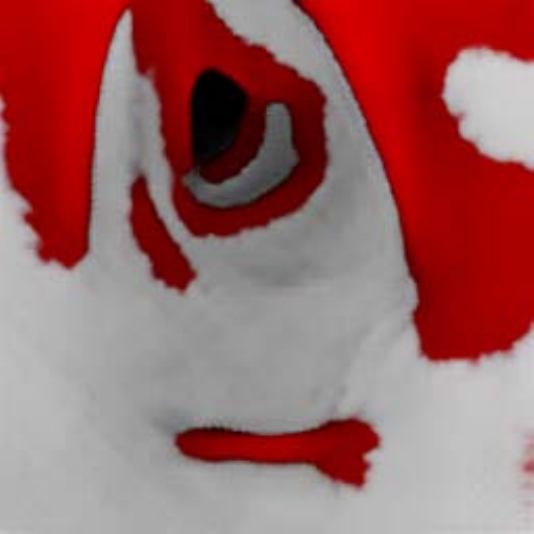}&
\includegraphics[width=0.11\textwidth]{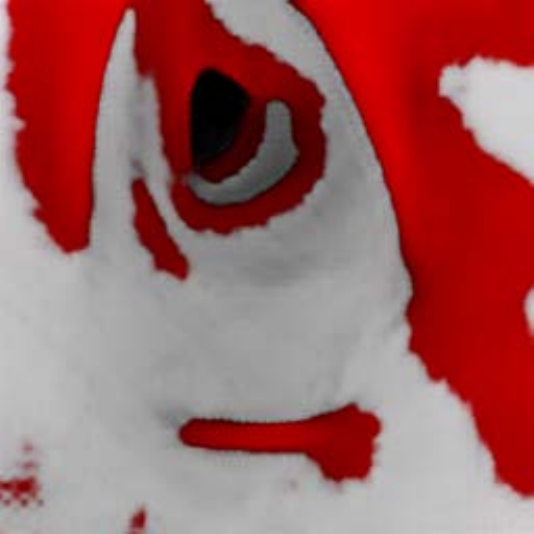}&
\includegraphics[width=0.11\textwidth]{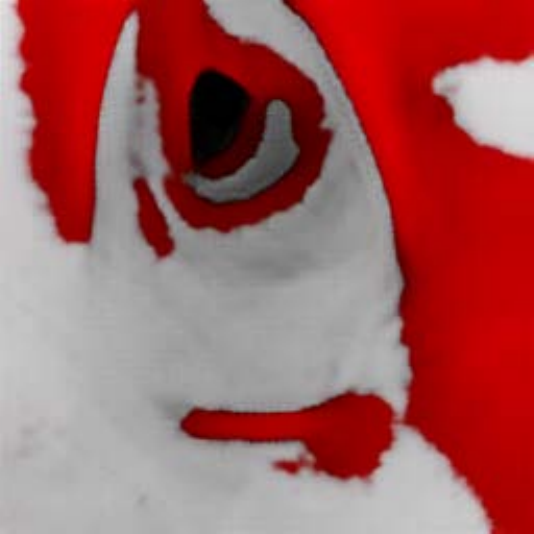}&
\includegraphics[width=0.11\textwidth]{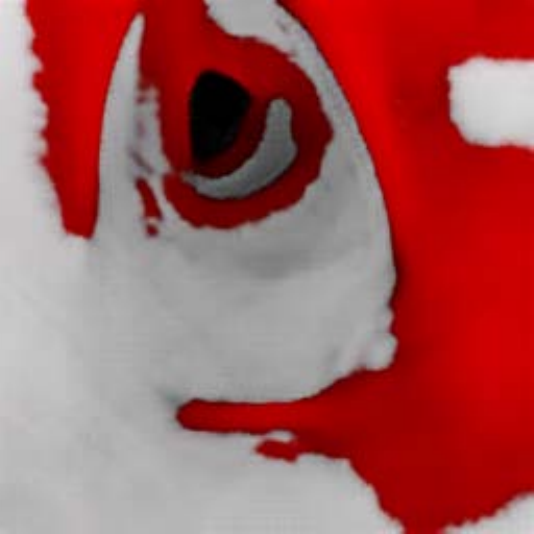}&
\includegraphics[width=0.11\textwidth]{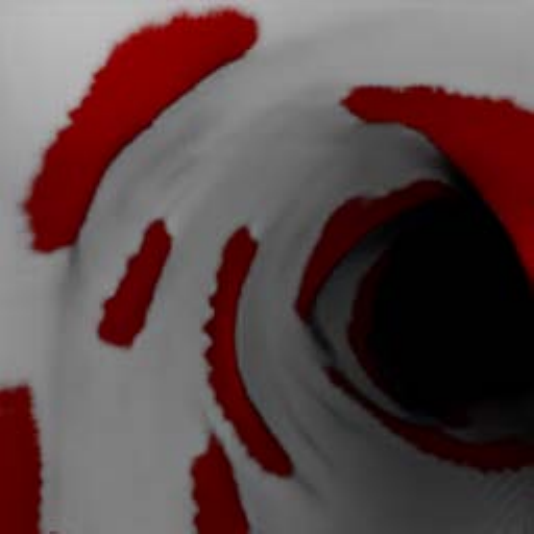}&
\includegraphics[width=0.11\textwidth]{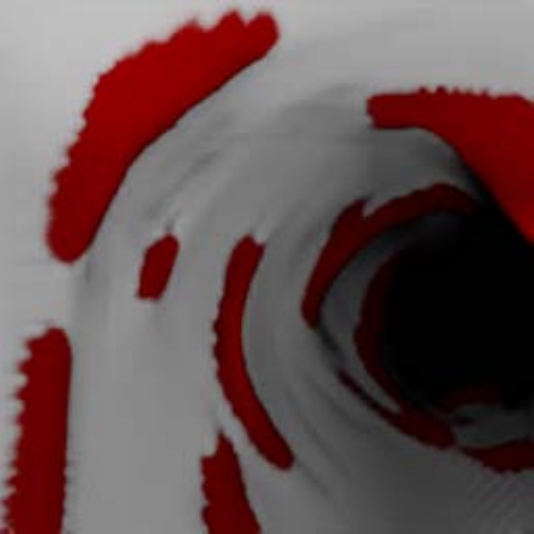}&
\includegraphics[width=0.11\textwidth]{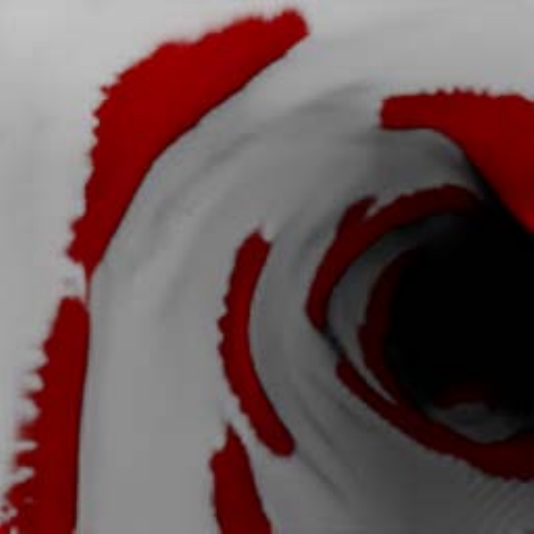}&
\includegraphics[width=0.11\textwidth]{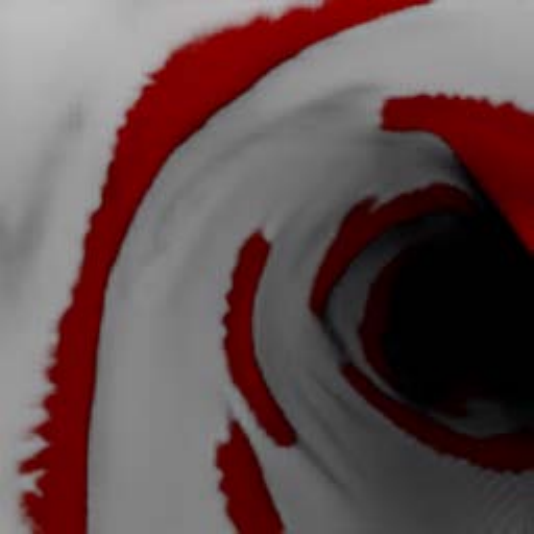}\\

\rotatebox{90}{\rlap{\small ~~$\lambda = 1$}}&
\includegraphics[width=0.11\textwidth]{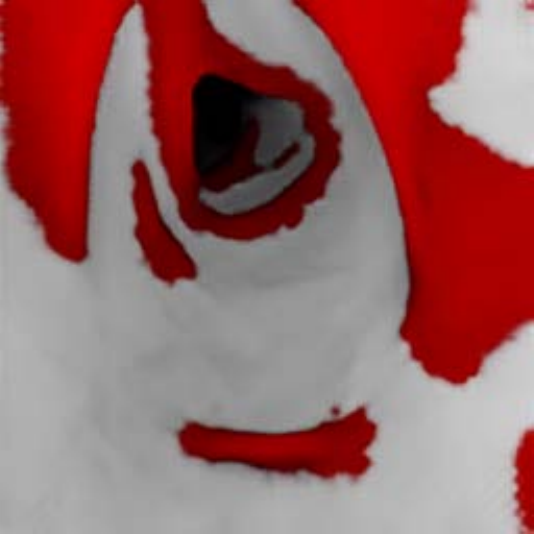}&
\includegraphics[width=0.11\textwidth]{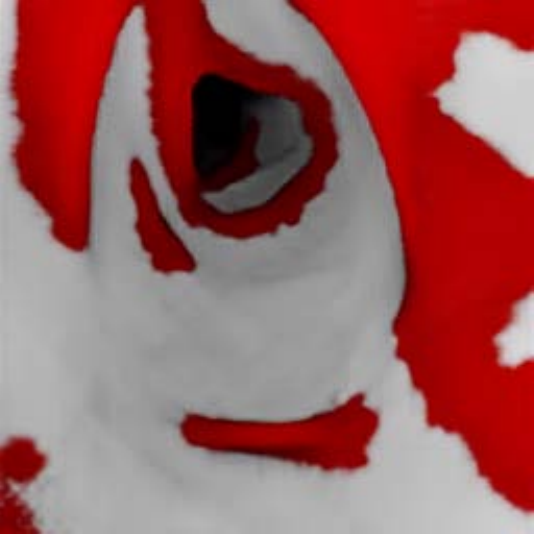}&
\includegraphics[width=0.11\textwidth]{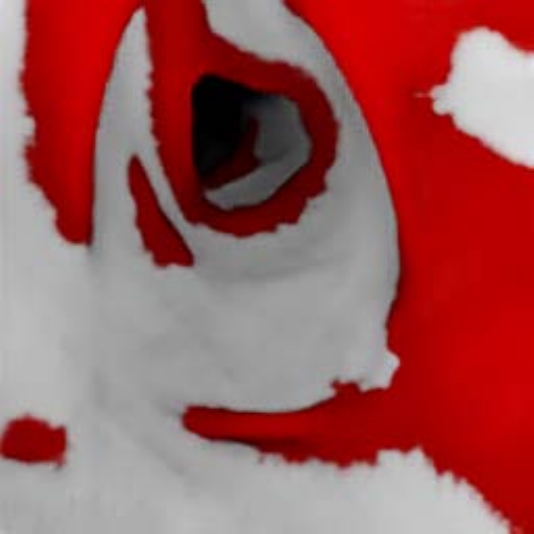}&
\includegraphics[width=0.11\textwidth]{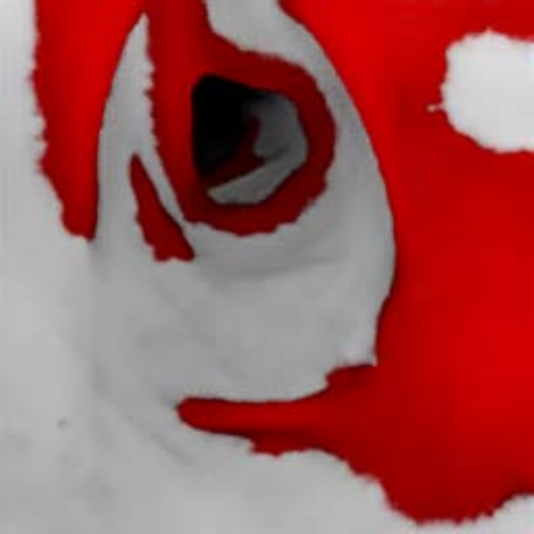}&
\includegraphics[width=0.11\textwidth]{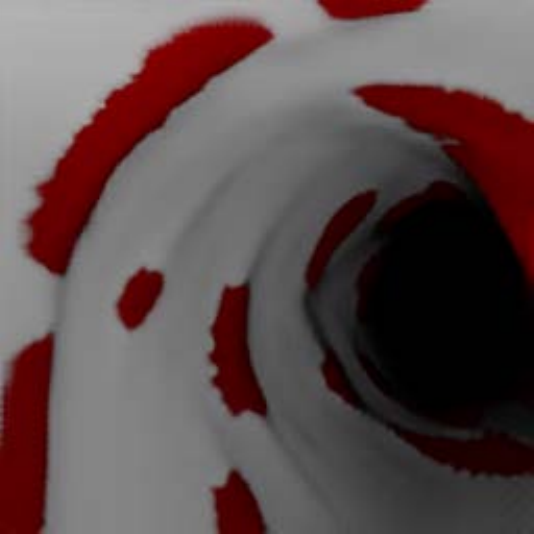}&
\includegraphics[width=0.11\textwidth]{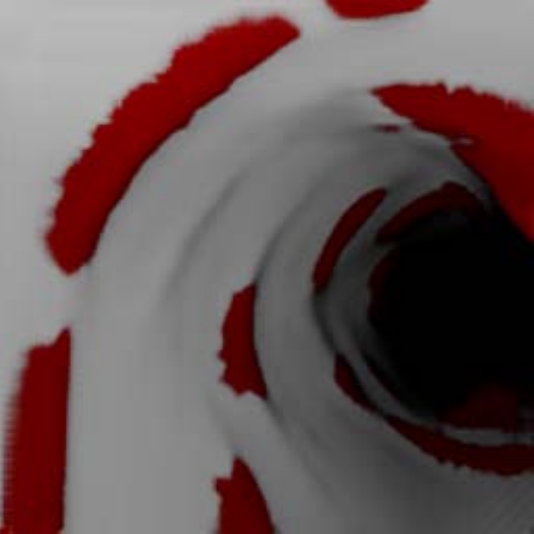}&
\includegraphics[width=0.11\textwidth]{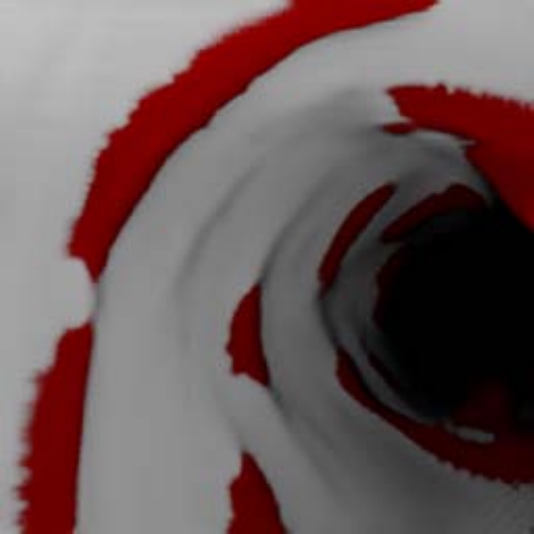}&
\includegraphics[width=0.11\textwidth]{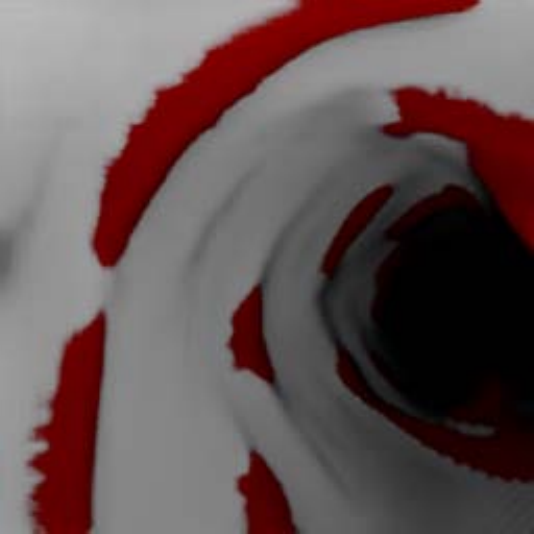}\\

\rotatebox{90}{\rlap{\small ~~$\lambda = 5$}}&
\includegraphics[width=0.11\textwidth]{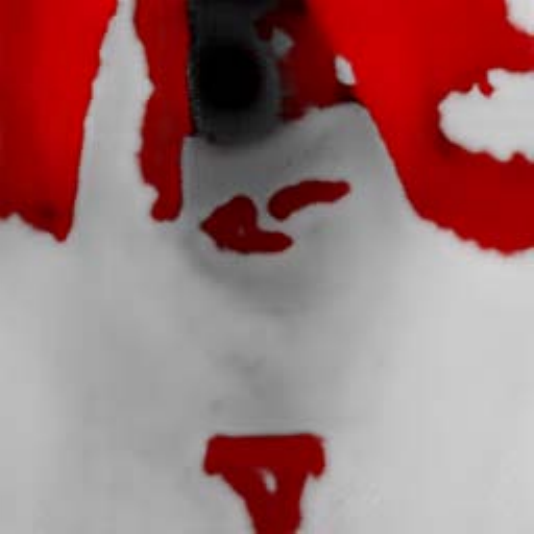}&
\includegraphics[width=0.11\textwidth]{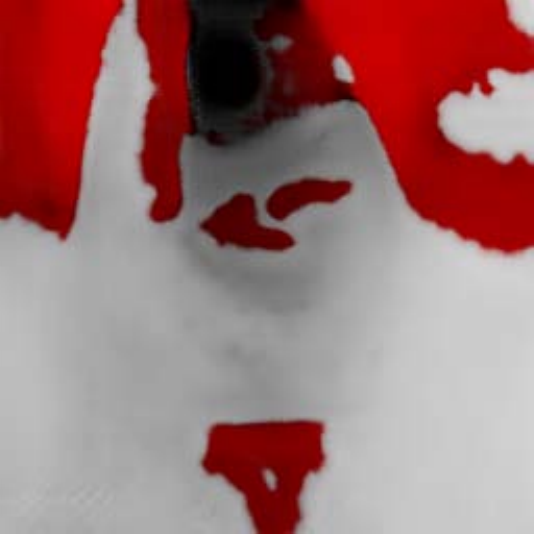}&
\includegraphics[width=0.11\textwidth]{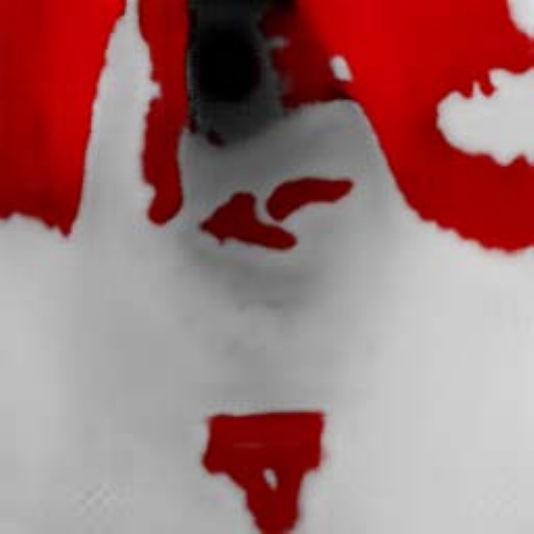}&
\includegraphics[width=0.11\textwidth]{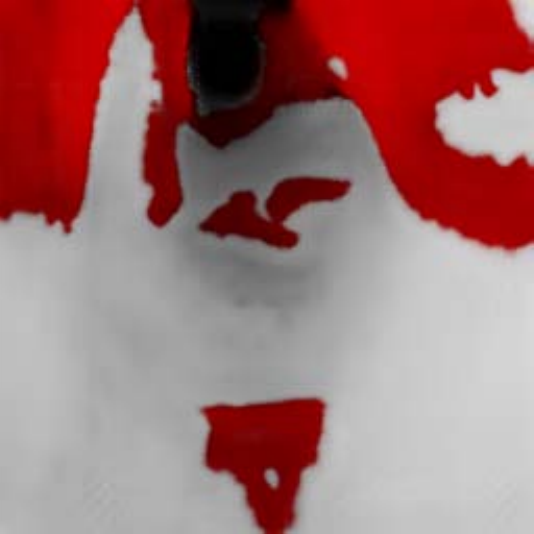}&
\includegraphics[width=0.11\textwidth]{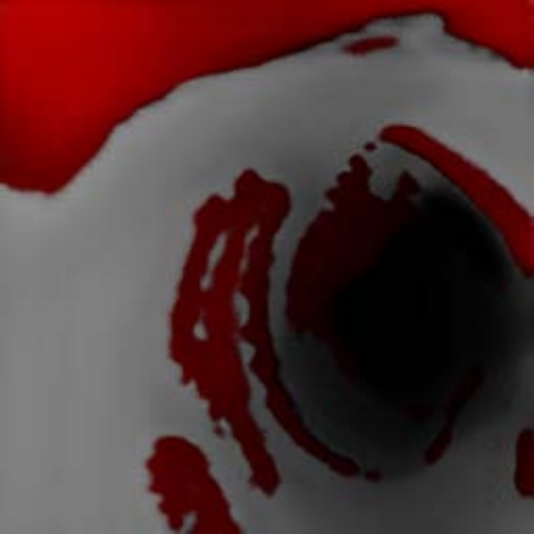}&
\includegraphics[width=0.11\textwidth]{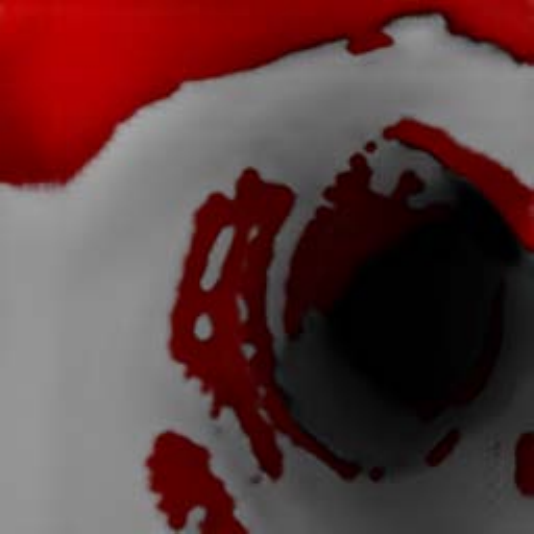}&
\includegraphics[width=0.11\textwidth]{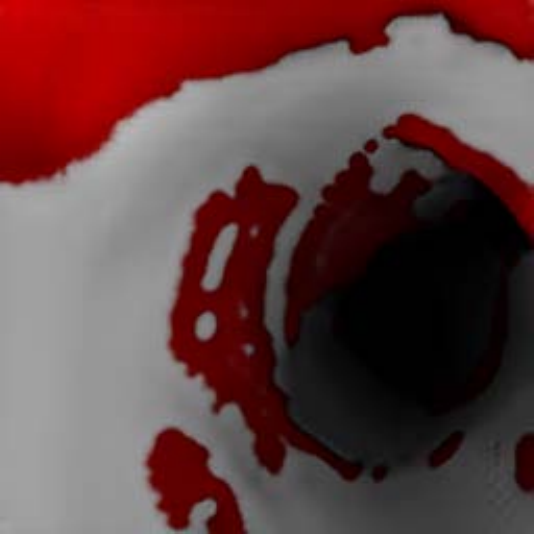}&
\includegraphics[width=0.11\textwidth]{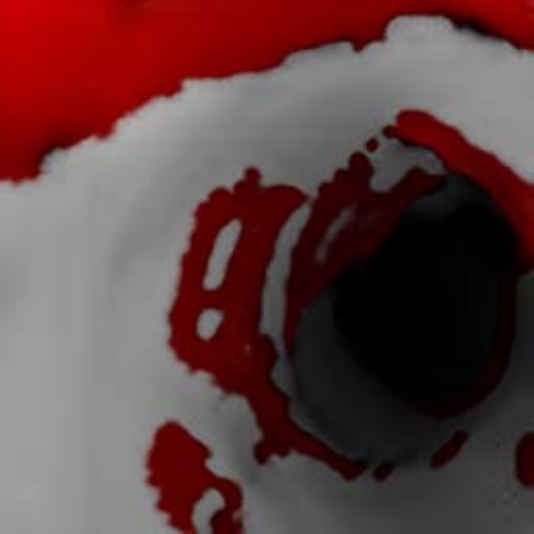}\\

\end{tabular}
\caption{Results for varying temporal weights ($\lambda$). The first row is the input and the second row shows $\lambda= 0.2$. As $\lambda$ decreases RT-GAN's is more faithful to FoldIt. The next row shows $\lambda=1$ where there is a balance between temporal and the frame losses. The last row shows $\lambda= 5$. Here the annotation shapes tend to remain consistent between frames. Full videos are found in the \textbf{supplementary video}.}
\label{fig:weight_comp}
\end{center}
\end{figure*}

\section{Results and Discussion}


\begin{table}[t]
\setlength{\tabcolsep}{2pt}
\caption{Number of learnable parameters (in millions) and training time per epoch for RecycleGAN \cite{bansal2018recycle}, TempCycleGAN\cite{engelhardt2018improving}, OfGAN \cite{xu2020ofgan}, FoldIt \cite{mathew2021foldit}, CLTS-GAN \cite{mathew2022clts}, and RT-GAN (ours). Models were trained on NVIDIA Quadro RTX 6000 GPU.}
\label{tab:params}
\centering

\begin{tabular}{|c|c|c|}
\hline
    & Learnable Parameters & Training Time \\
    \hline
    RecycleGAN \cite{bansal2018recycle} & 137.11  & $\sim$ 742 s \\
    TempCycleGAN \cite{engelhardt2018improving} & 46.65  & $\sim$ 836 s \\
    OfGAN \cite{xu2020ofgan} & 142.23 & $\sim$ 947 s \\
    FoldIt \cite{mathew2021foldit} & 82.14 & $\sim$ 700 s \\ 
    CLTS-GAN \cite{mathew2022clts} & 55.15 & $\sim$ 857 s \\
    \textbf{RT-GAN (Ours)} & \textbf{25.22} & \textbf{$\sim$ 394 s} \\
\hline          
\end{tabular}
\end{table}

\begin{table}[t!]
    \caption{Quantitative results on a synthetic colon dataset \cite{mathew2021foldit} with two textures and ground truth fold annotations. The consistency column indicates the frame-based geometric consistency of the model despite different textures as described by Mathew et al.\cite{mathew2020augmenting}. These sequences are shown in the \textbf{supplementary video}.}
    \label{tab:quant}
    \centering
    \setlength{\tabcolsep}{2pt}
    \begin{tabular}{|c||c|c|c|}
     \hline    
       & Text 1 (IoU/DICE) & Text 2 (IoU/DICE) & Consistency (IoU/DICE)\\
       \hline
       RecycleGAN & $0.34$/$0.21$ & $0.33$/$0.20$ &  $0.76$/$0.63$\\
       FoldIt & $0.47$/$0.31$ & $0.50$/$0.33$ & $0.77$/$0.64$\\
       RT-GAN  & $\mathbf{0.55}$/$\mathbf{0.39}$ & $\mathbf{0.54 }$/$\mathbf{0.38}$ & $\mathbf{0.81}$/$\mathbf{0.69}$\\
       \hline
    \end{tabular}
\end{table}

The training time and memory usage of RT-GAN is analyzed in Table \ref{tab:params}. 
RT-GAN reduces the number of learnable parameters by a factor of 5 while decreasing the training time by half when compared with RecycleGAN and OfGAN. Compared to TempCycleGAN, a video domain translation model with a minimal amount of image generators, RT-GAN reduces the number of learnable parameters and training time by a factor of 2. FoldIt, a frame-based model for haustral fold segmentation, uses fewer resources than RecycleGAN as it deals with individual frames. RT-GAN still requires lesser resources than FoldIt because it only learns one direction of translation while FoldIt learns four \cite{mathew2021foldit}. CLTS-GAN \cite{mathew2022clts} only learns two directions of translation, so RT-GAN reduces the learnable parameters in half. When training RT-GAN, the hardware requirements are capped by the frame-based model since RT-GAN requires lesser resources.

\begin{figure*}[t!]
\begin{center}
\setlength{\tabcolsep}{2pt}
\begin{tabular}{ccccc|cccc}

\rotatebox{90}{\rlap{\small ~~Input}}&
\includegraphics[width=0.11\textwidth]{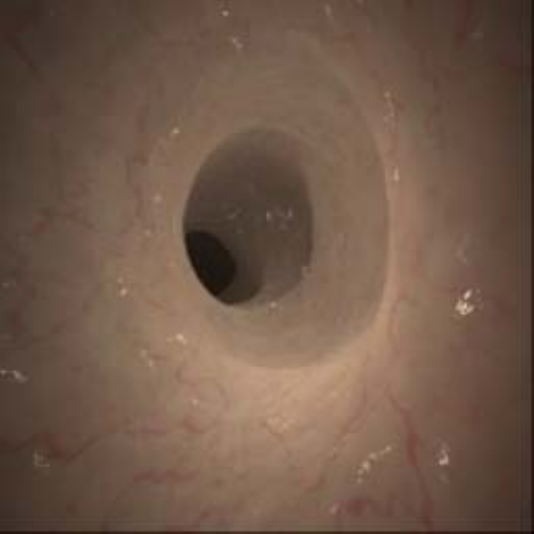}&
\includegraphics[width=0.11\textwidth]{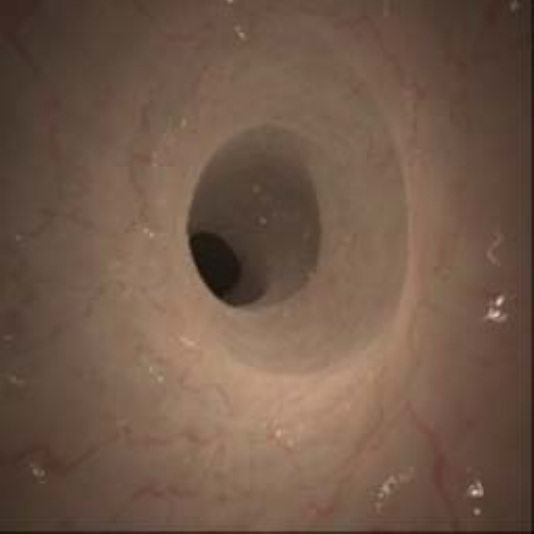}&
\includegraphics[width=0.11\textwidth]{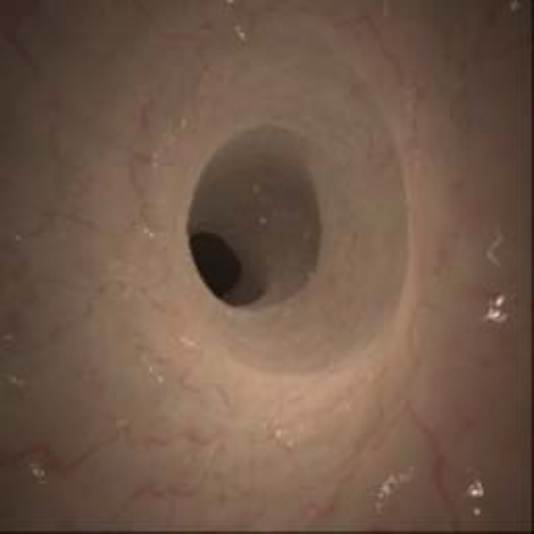}&
\includegraphics[width=0.11\textwidth]{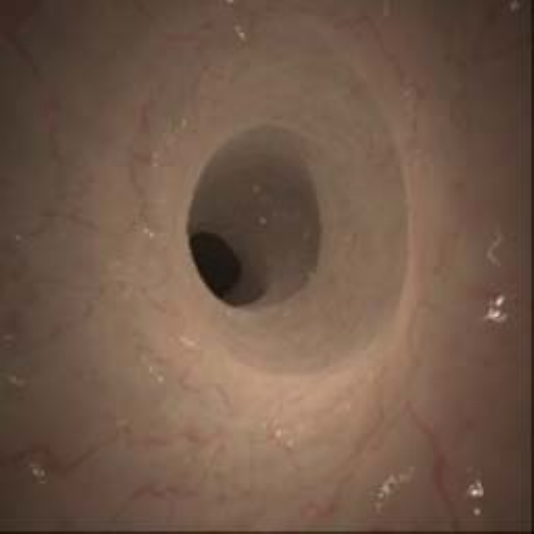}&
\includegraphics[width=0.11\textwidth]{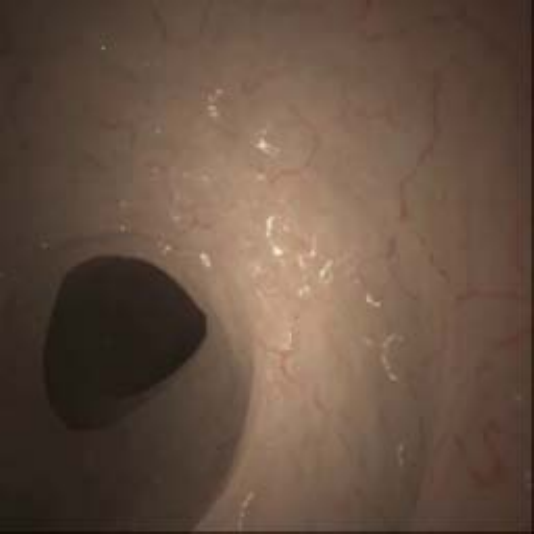}&
\includegraphics[width=0.11\textwidth]{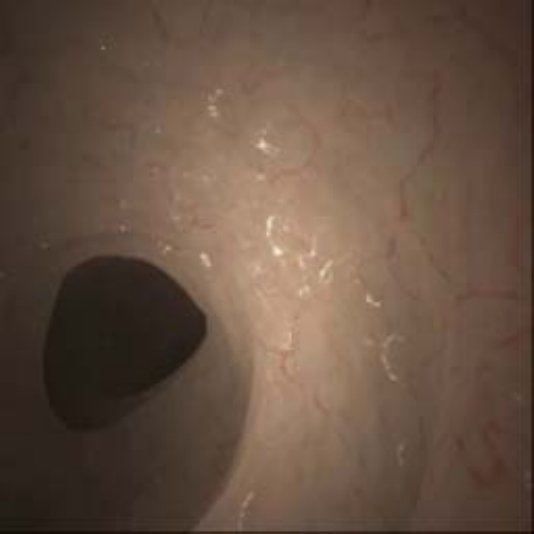}&
\includegraphics[width=0.11\textwidth]{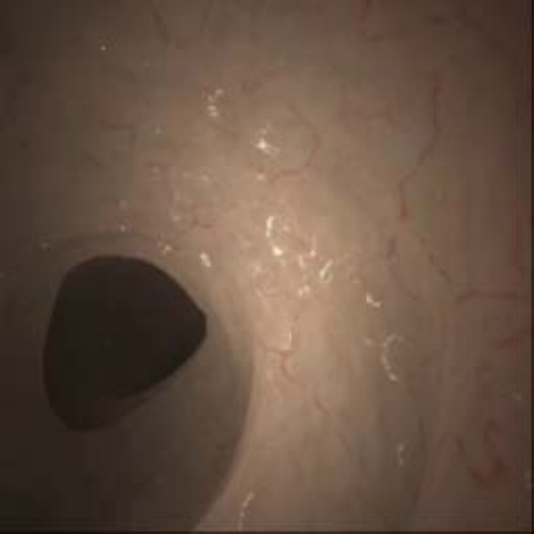}&
\includegraphics[width=0.11\textwidth]{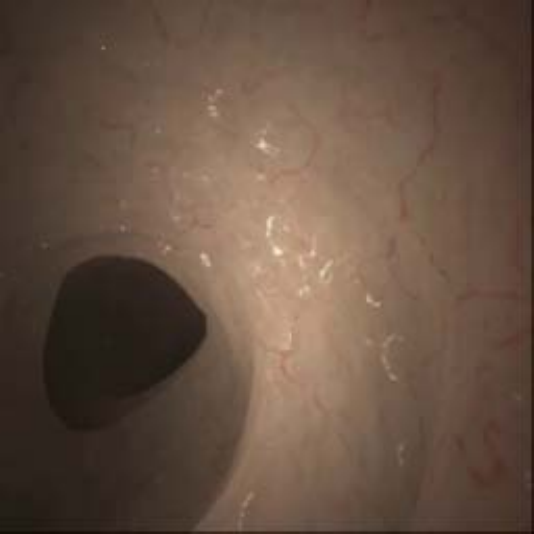}\\

\rotatebox{90}{\rlap{\tiny ~CLTS-GAN}}&
\includegraphics[width=0.11\textwidth]{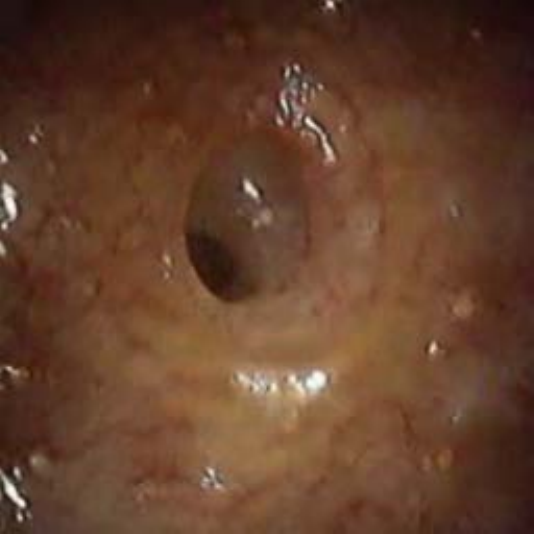}&
\includegraphics[width=0.11\textwidth]{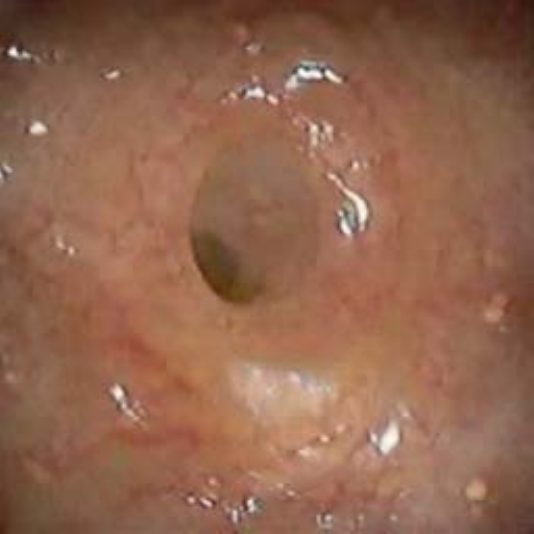}&
\includegraphics[width=0.11\textwidth]{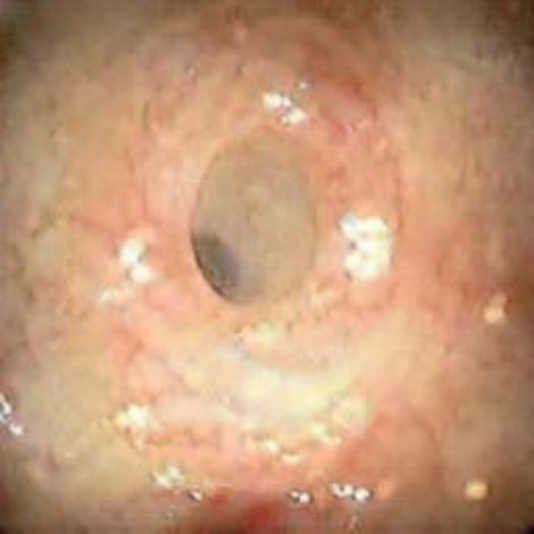}&
\includegraphics[width=0.11\textwidth]{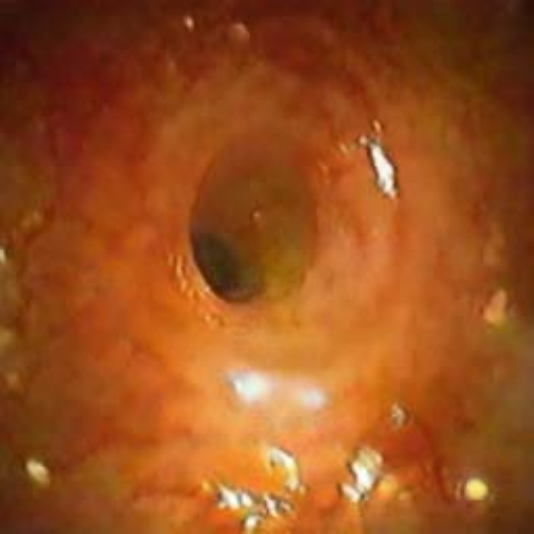}&
\includegraphics[width=0.11\textwidth]{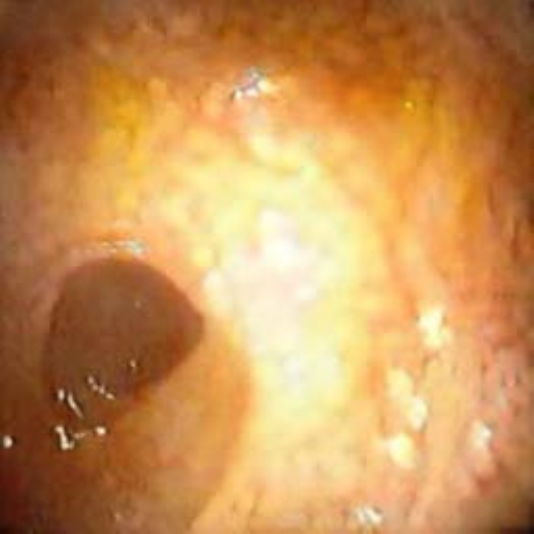}&
\includegraphics[width=0.11\textwidth]{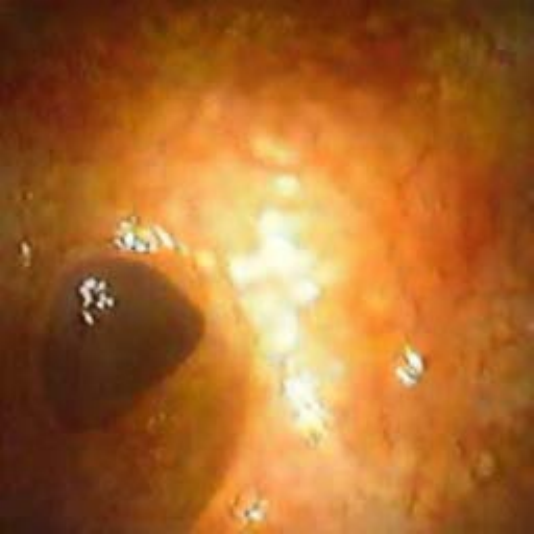}&
\includegraphics[width=0.11\textwidth]{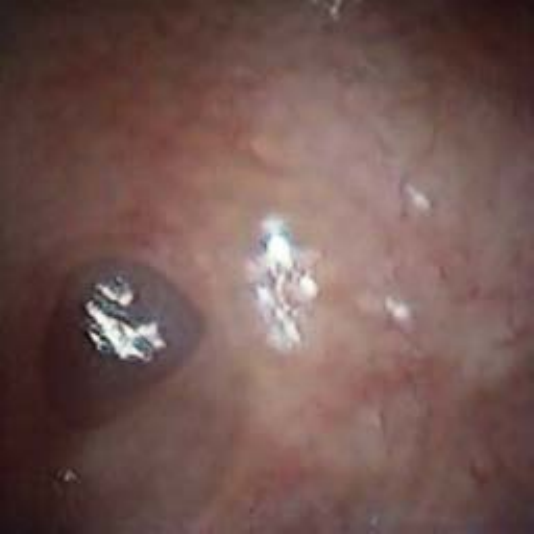}&
\includegraphics[width=0.11\textwidth]{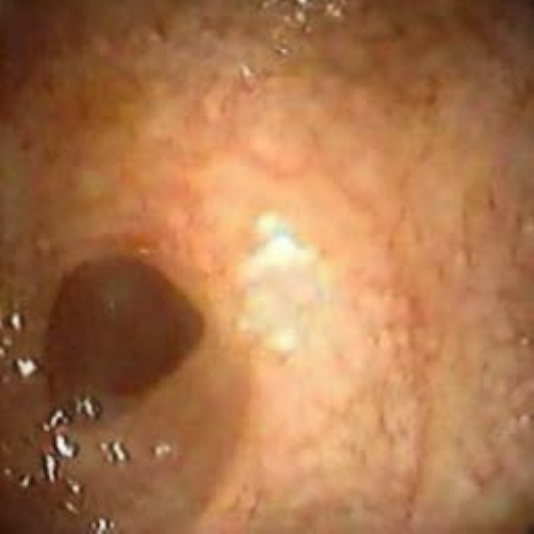}\\

\rotatebox{90}{\rlap{\tiny ~OfGAN}}&
\includegraphics[width=0.11\textwidth]{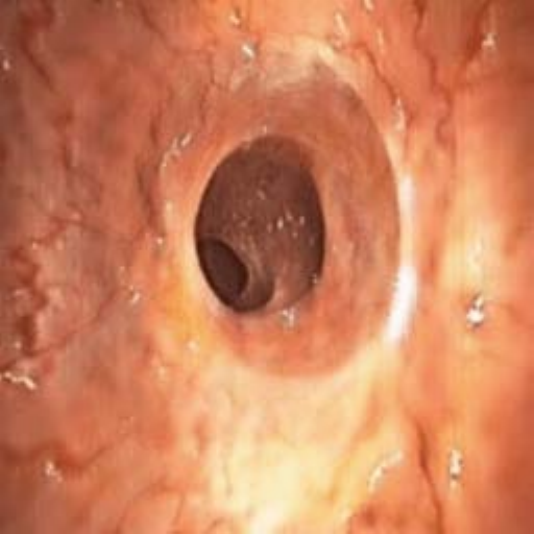}&
\includegraphics[width=0.11\textwidth]{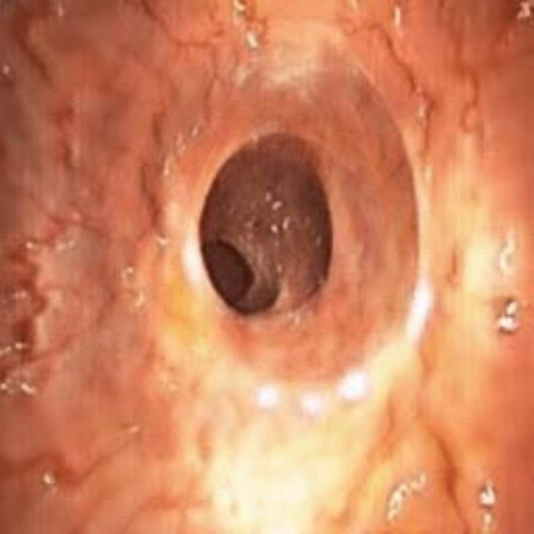}&
\includegraphics[width=0.11\textwidth]{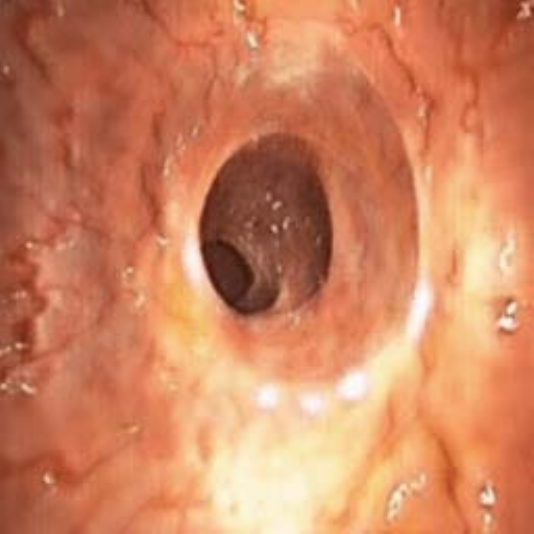}&
\includegraphics[width=0.11\textwidth]{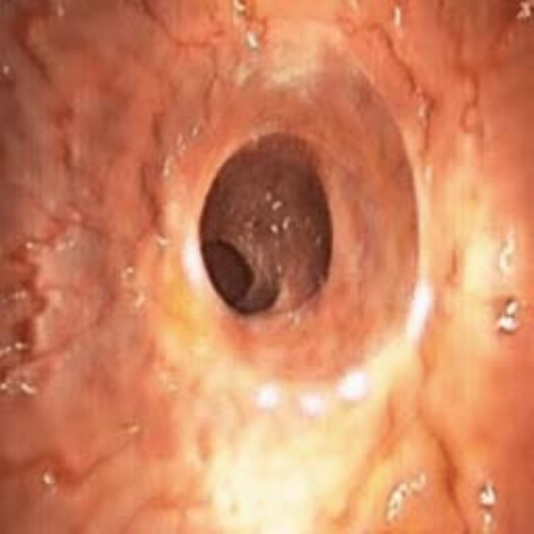}&
\includegraphics[width=0.11\textwidth]{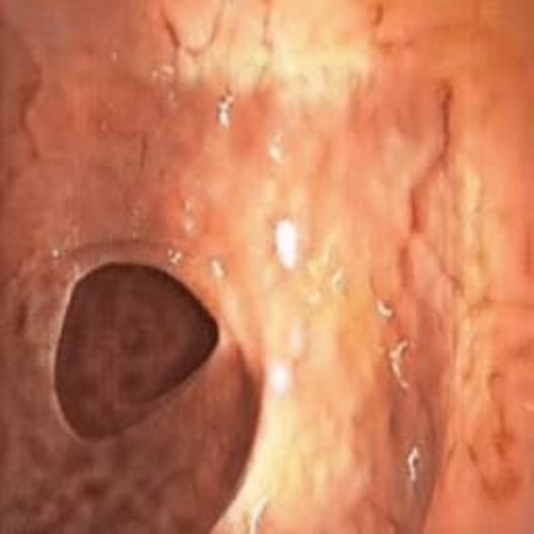}&
\includegraphics[width=0.11\textwidth]{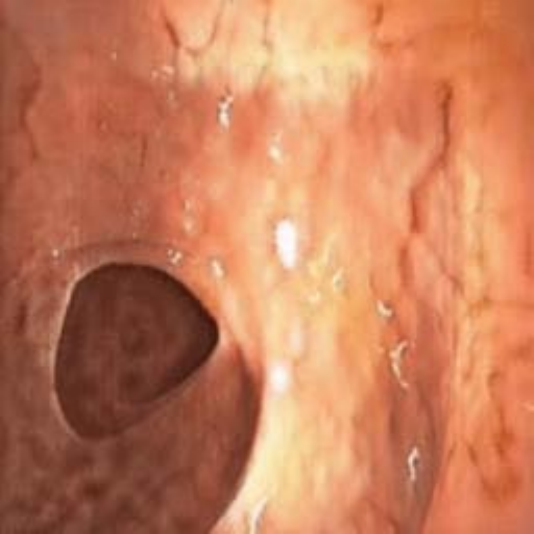}&
\includegraphics[width=0.11\textwidth]{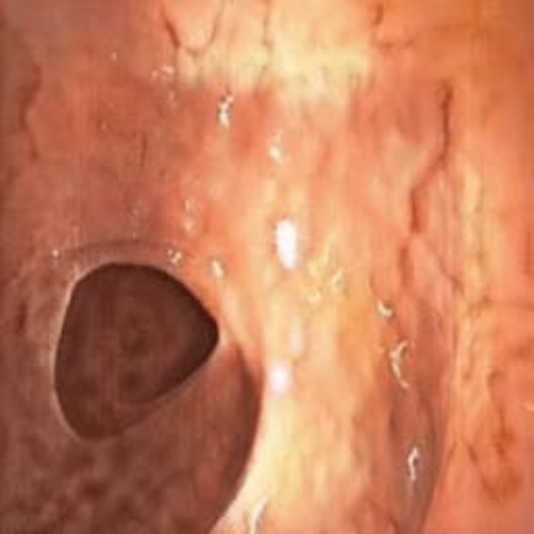}&
\includegraphics[width=0.11\textwidth]{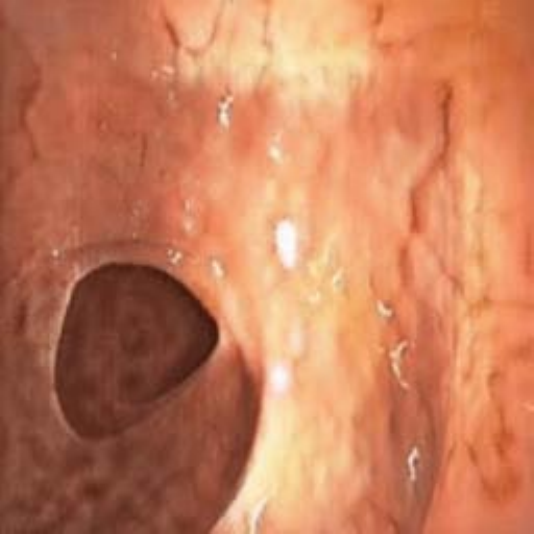}\\

\rotatebox{90}{\rlap{\tiny ~RT-GAN}}&
\includegraphics[width=0.11\textwidth]{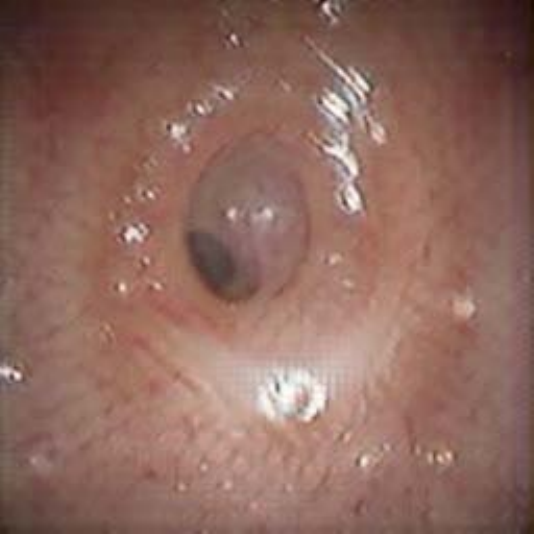}&
\includegraphics[width=0.11\textwidth]{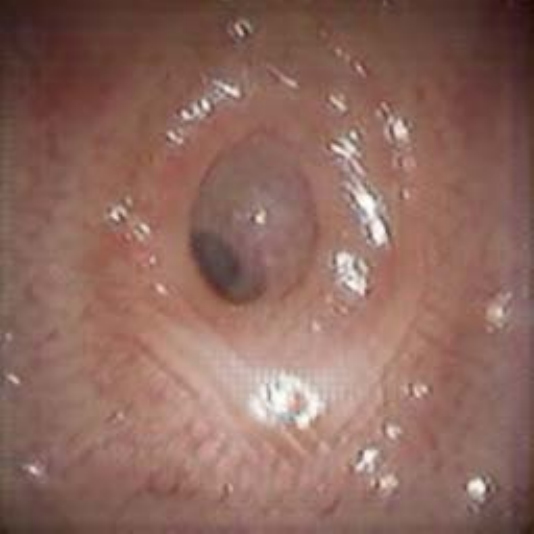}&
\includegraphics[width=0.11\textwidth]{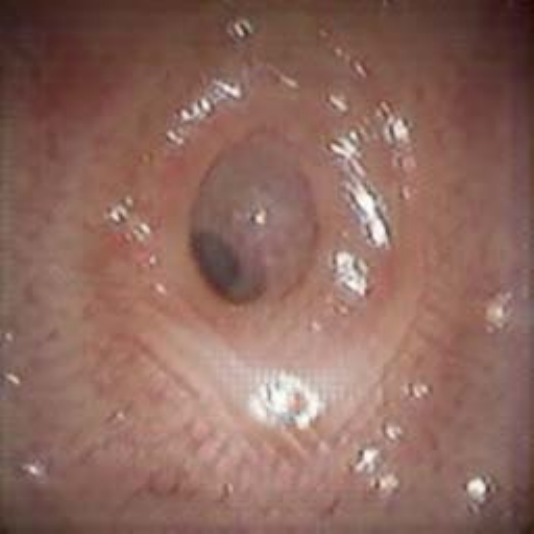}&
\includegraphics[width=0.11\textwidth]{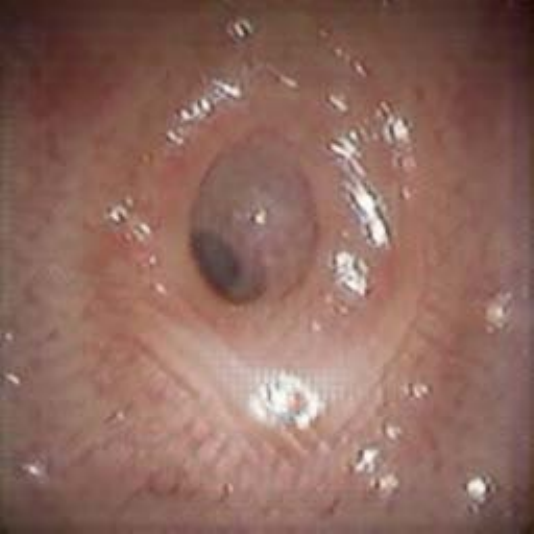}&
\includegraphics[width=0.11\textwidth]{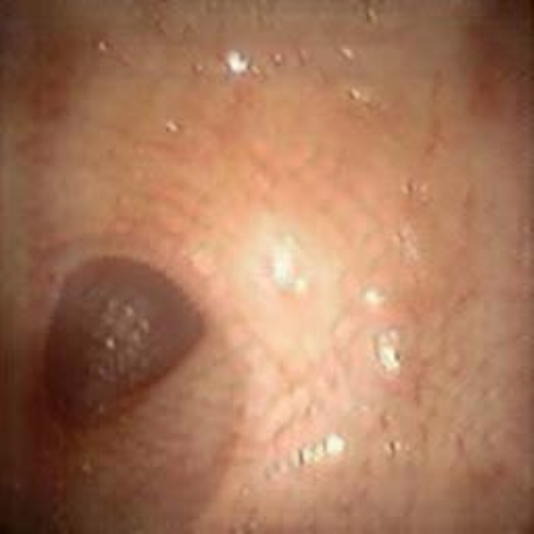}&
\includegraphics[width=0.11\textwidth]{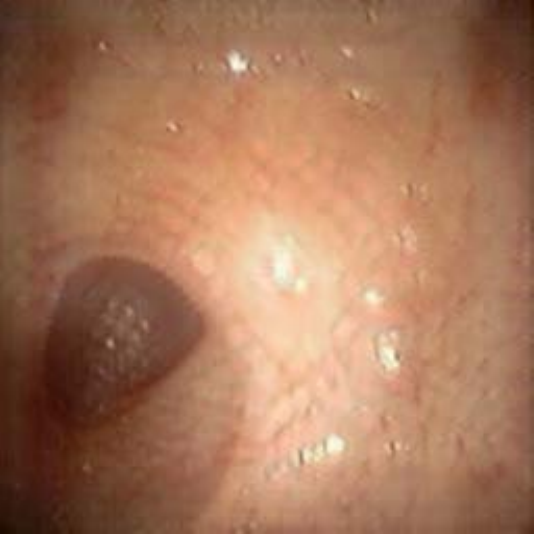}&
\includegraphics[width=0.11\textwidth]{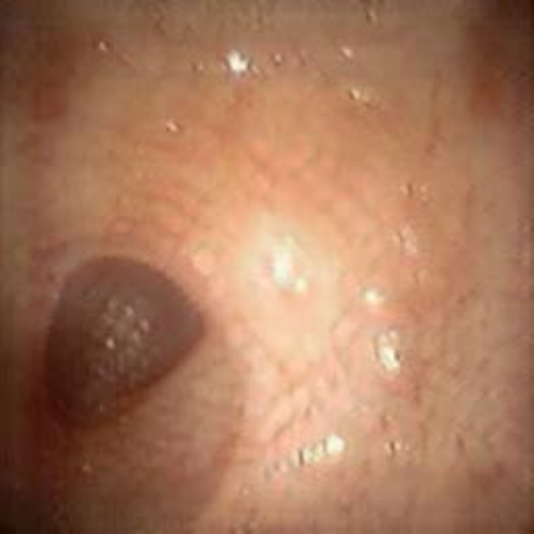}&
\includegraphics[width=0.11\textwidth]{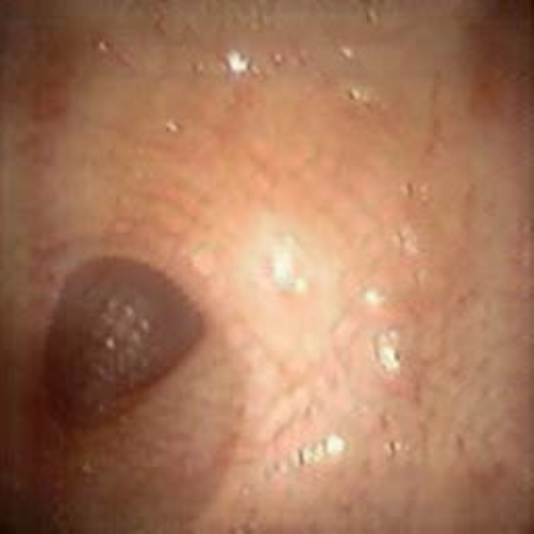}\\

\end{tabular}
\caption{Results for RT-GAN trained on CLTS-GAN. The top portion shows results on rendered mesh frames. CLTS-GAN's results change drastically over time. RT-GAN builds off CLTS-GAN to provide consistent specular and texture between frames. The bottom half shows results using OfGAN's input video, which embeds texture and specular information. CLTS-GAN adds more intricate specular reflections and textures, and RT-GAN inherits this property. OfGAN relies on the embedded texture and specular to produce its output. Full videos are in the \textbf{supplementary video}.}
\label{fig:OC_gen}
\end{center}
\end{figure*}

To test the effectiveness of RT-GAN in fold segmentation context (indicative of the total missed surface during colonoscopy), we added RT-GAN on top of FoldIt haustral fold frame-based model \cite{mathew2021foldit}. In Figure \ref{fig:recycle_comp}, we compare RT-GAN, FoldIt, TempCycleGAN, and RecycleGAN results on public video sequences from Ma et al. \cite{ma2019real}. RecycleGAN has many variants, however sifting through all the variants and applying task-specific components requires great effort on part of the end users. We chose RecycleGAN for comparisons since it has all the base temporal components seen in the more advanced variants and is not task-specific. As shown in supplementary video, FoldIt and RecycleGAN both had jittery results and FoldIt occasionally smooths out the deeper parts of the endolumen. In contrast, RecycleGAN translated these deeper endolumen parts as folds since it does not contain any task-specific modules or losses. RT-GAN utilizes the task-specific modules from FoldIt while providing temporal consistency. TempCycleGAN is more consistent, however, similar to RecycleGAN it doesn't have the task specific additions and it fails to accommodate the deeper portions of the endoluminal view. Complete videos sequences are shown in the \textbf{supplement}.

For quantitative analysis, synthetic colon dataset with ground truth annotations was used \cite{mathew2021foldit}. Table \ref{tab:quant} shows that RT-GAN's additional temporal consistency provided improvement on the IoU and DICE scores for both textures. RT-GAN is also more consistent than the other models despite different textures. Additionally, the optical flow can be compared in the input sequences and output sequences as done by Rivoir et al. \cite{rivoir2021long}. The mean difference between the input optical flow and output optical flow on our textured colons for RecycleGAN, FoldIt, and RT-GAN are 2.4788, 0.9021, and 0.8479, respectively. This indicates that RT-GAN can better capture the motion between frames when compared with other models like RecycleGAN and FoldIt. The synthetic colon results can be found in the supplementary video. In Figure \ref{fig:weight_comp}, the $\lambda$ parameter to control temporal consistency is shown. When $\lambda$ is set to a lower value, it tries to be more faithful to FoldIt. As $\lambda$ is increased, RT-GAN makes the annotations smoother so it looks more temporally consistent.

We also evaluated RT-GAN on real colonoscopy video generation/simulation using the frame-based CLTS-GAN model \cite{mathew2022clts}. CLTS-GAN creates colonoscopy frames with different colors, lighting, textures, and specular reflections using noise parameters. For real colonoscopy video generation/simulation, RT-GAN was trained for 200 epochs on 1800 frame triplets of colonoscopy video and 3D renderings of the colon using virtual colonoscopy from \cite{mathew2022clts}. The results of real colonoscopy video generation from synthetic sequences are shown in Figure \ref{fig:OC_gen}. The top half shows video generation from virtual colonoscopy renderings. CLTS-GAN's use of noise parameters allows it to generate drastically different output across frames. RT-GAN is much smoother and the specular reflections and textures are consistent; in the \textbf{supplementary video}, the overall color and lighting changes over time since RT-GAN only looks at the previous frame (and doesn't have a longer-term memory, an issue we will resolve in the future). The bottom portion of Figure \ref{fig:OC_gen} compares (RT-GAN + CLTS-GAN) with OfGAN \cite{xu2020ofgan}. OfGAN is confined to creating textures and specular reflections that are embedded in its input video. In contrast, CLTS-GAN adds additional texture and specular reflections but lacks temporal consistency. RT-GAN uses CLTS-GAN's texture and specular information and adds temporal consistency on top of it. Complete video results are shown in the \textbf{supplement}.


\noindent 
\textbf{Limitations.} The first is the lack of long term memory since RT-GAN only receives information from the previous frame. Incorporating transformers or State Space Models \cite{fuoli2023efficient,neimark2021video,ullah2017action,ryali2023hiera} could mitigate this issue. Additionally, RT-GAN can inherit some of the limitations of its frame based model, e.g. FoldIt cannot handle frame occlusion and hence both FoldIt and RT-GAN can hallucinate endoluminal view for occluded frames. In the future, we will explore the application of RT-GAN to other endoscopy procedures, such as cystoscopy, bronchoscopy, and naseopharyngoscopy.

\begin{credits}
\subsubsection{\ackname} This project was supported by NIH R37CA295658, MSK Cancer Center Support Grant/Core Grant (P30 CA008748), and NSF grants CNS1650499, OAC1919752, and ICER1940302.

\subsubsection{\discintname}
Dr. Nadeem has received speaker honorarium from Roche Tissue Diagnostics which is not related to this work. He also serves on MONAI (Medical Open Network for Artificial Intelligence) consortium advisory board, unrelated to this work. Alvin C. Goh receives research support from Intuitive Surgical. Other authors do not declare any competing interests. 
\end{credits}

%
%
%

\begin{thebibliography}{10}
\providecommand{\url}[1]{\texttt{#1}}
\providecommand{\urlprefix}{URL }
\providecommand{\doi}[1]{https://doi.org/#1}

\bibitem{bansal2018recycle}
Bansal, A., Ma, S., Ramanan, D., Sheikh, Y.: Recycle-gan: Unsupervised video
  retargeting. Proceedings of the European Conference on Computer Vision (ECCV)
  pp. 119--135 (2018)

\bibitem{bashkirova2018unsupervised}
Bashkirova, D., Usman, B., Saenko, K.: Unsupervised video-to-video translation.
  arXiv preprint arXiv:1806.03698  (2018)

\bibitem{daher2023cyclesttn}
Daher, R., Barbed, O.L., Murillo, A.C., Vasconcelos, F., Stoyanov, D.:
  Cyclesttn: A learning-based temporal model for specular augmentation in
  endoscopy. International Conference on Medical Image Computing and
  Computer-Assisted Intervention pp. 570--580 (2023)

\bibitem{engelhardt2018improving}
Engelhardt, S., De~Simone, R., Full, P.M., Karck, M., Wolf, I.: Improving
  surgical training phantoms by hyperrealism: deep unpaired image-to-image
  translation from real surgeries. International Conference on Medical Image
  Computing and Computer-Assisted Intervention (MICCAI) pp. 747--755 (2018)

\bibitem{fuoli2023efficient}
Fuoli, D., Huang, Z., Paudel, D.P., Van~Gool, L., Timofte, R.: An efficient
  recurrent adversarial framework for unsupervised real-time video enhancement.
  International Journal of Computer Vision pp. 1--18 (2023)

\bibitem{incetan2021vr}
{\.I}ncetan, K., Celik, I.O., Obeid, A., Gokceler, G.I., Ozyoruk, K.B.,
  Almalioglu, Y., Chen, R.J., Mahmood, F., Gilbert, H., Durr, N.J., Turana, M.:
  {VR-Caps}: A virtual environment for capsule endoscopy. Medical Image
  Analysis p. 101990 (2021)

\bibitem{ma2019real}
Ma, R., Wang, R., Pizer, S., Rosenman, J., McGill, S.K., Frahm, J.M.: Real-time
  {3D} reconstruction of colonoscopic surfaces for determining missing regions.
  International Conference on Medical Image Computing and Computer-Assisted
  Intervention (MICCAI) pp. 573--582 (2019)

\bibitem{mahmood2018unsupervised}
Mahmood, F., Chen, R., Durr, N.J.: Unsupervised reverse domain adaptation for
  synthetic medical images via adversarial training. IEEE Transactions on
  Medical Imaging  \textbf{37}(12),  2572--2581 (2018)

\bibitem{mathew2021foldit}
Mathew, S., Nadeem, S., Kaufman, A.: Foldit: Haustral folds detection and
  segmentation in colonoscopy videos. International Conference on Medical Image
  Computing and Computer-Assisted Intervention (MICCAI) pp. 221--230 (2021)

\bibitem{mathew2022clts}
Mathew, S., Nadeem, S., Kaufman, A.: {CLTS-GAN:}
  color-lighting-texture-specular reflection augmentation for colonoscopy.
  International Conference on Medical Image Computing and Computer-Assisted
  Intervention (MICCAI) pp. 519--529 (2022)

\bibitem{mathew2020augmenting}
Mathew, S., Nadeem, S., Kumari, S., Kaufman, A.: Augmenting colonoscopy using
  extended and directional {CycleGAN} for lossy image translation. Proceedings
  of the IEEE/CVF Conference on Computer Vision and Pattern Recognition (CVPR)
  pp. 4696--4705 (2020)

\bibitem{nadeem2016computer}
Nadeem, S., Kaufman, A.: Computer-aided detection of polyps in optical
  colonoscopy images. SPIE Medical Imaging  \textbf{9785},  978525 (2016)

\bibitem{neimark2021video}
Neimark, D., Bar, O., Zohar, M., Asselmann, D.: Video transformer network.
  Proceedings of the IEEE/CVF International Conference on Computer Vision
  (ICCV) pp. 3163--3172 (2021)

\bibitem{rivoir2021long}
Rivoir, D., Pfeiffer, M., Docea, R., Kolbinger, F., Riediger, C., Weitz, J.,
  Speidel, S.: Long-term temporally consistent unpaired video translation from
  simulated surgical {3D} data. Proceedings of the IEEE/CVF International
  Conference on Computer Vision (ICCV) pp. 3343--3353 (2021)

\bibitem{ryali2023hiera}
Ryali, C., Hu, Y.T., Bolya, D., Wei, C., Fan, H., Huang, P.Y., Aggarwal, V.,
  Chowdhury, A., Poursaeed, O., Hoffman, J., et~al.: Hiera: A hierarchical
  vision transformer without the bells-and-whistles. Proceedings of the
  International Conference on Machine Learning (ICML)  (2023)

\bibitem{ullah2017action}
Ullah, A., Ahmad, J., Muhammad, K., Sajjad, M., Baik, S.W.: Action recognition
  in video sequences using deep bi-directional lstm with cnn features. IEEE
  Access  \textbf{6},  1155--1166 (2017)

\bibitem{xu2020ofgan}
Xu, J., Anwar, S., Barnes, N., Grimpen, F., Salvado, O., Anderson, S., Armin,
  M.A.: Ofgan: Realistic rendition of synthetic colonoscopy videos.
  International Conference on Medical Image Computing and Computer-Assisted
  Intervention (MICCAI) pp. 732--741 (2020)

\end{thebibliography}

%




\end{document}